\documentclass{article} % For LaTeX2e
\usepackage{iclr2026_conference,times}

\usepackage{amsmath,amsfonts,bm}

\def\eqref#1{equation~\ref{#1}}
\def\1{\bm{1}}

\DeclareMathAlphabet{\mathsfit}{\encodingdefault}{\sfdefault}{m}{sl}
\SetMathAlphabet{\mathsfit}{bold}{\encodingdefault}{\sfdefault}{bx}{n}

\usepackage{booktabs}
\usepackage{hyperref}
\usepackage{url}
\usepackage{graphicx}
\usepackage{longtable}
\usepackage{paralist}
\usepackage{enumitem}
\usepackage[latin1]{inputenc}
\usepackage[normalem]{ulem}  % strikeout text
\usepackage{siunitx} % pretty quantities 23,456
\usepackage{pifont} % Checkmark crossmark
\usepackage{xcolor} % Checkmark crossmark
\usepackage{makecell}
\usepackage{wrapfig}
\usepackage{minted}
\usepackage{subcaption}

\newcommand{\withsup}[2]{#1\textsuperscript{\textcolor{green!70!black}{\scriptsize +#2\%}}}

\title{
\fontsize{16.9}{20}\selectfont
TOUCAN: Synthesizing 1.5M Tool-Agentic Data from Real-World MCP Environments}

\newcommand{\github}{\raisebox{-1.5pt}{\includegraphics[height=1em]{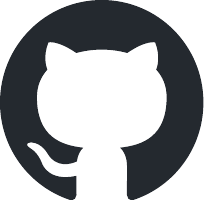}}}
\newcommand{\huggingface}{\raisebox{-1.5pt}{\includegraphics[height=1em]{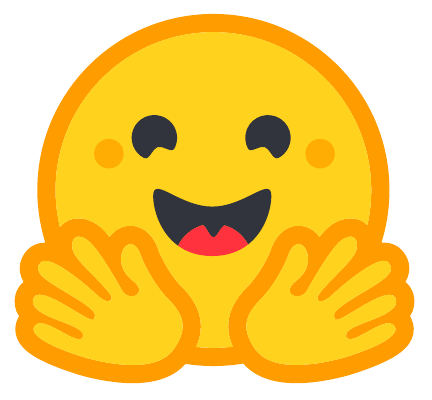}}}

\author{Zhangchen Xu$^\spadesuit$ \qquad Adriana Meza Soria $^\diamondsuit$ \qquad Shawn Tan$^\diamondsuit$
\qquad  Anurag Roy $^\diamondsuit$ ~\vspace{0.4em} \\
\textbf{Ashish Sunil Agrawal$^\diamondsuit$ \qquad  Radha Poovendran$^\spadesuit$ \qquad  Rameswar Panda$^\diamondsuit$}
~\vspace{0.5em}\\
$^\spadesuit$University of Washington \qquad
$^\diamondsuit$MIT-IBM Watson AI Lab \vspace{0.5em}
\\
{\small{\github{} {\texttt{\url{https://github.com/TheAgentArk/Toucan}}}}} \\
{\small{\huggingface{} {\texttt{\url{https://hf.co/datasets/Agent-Ark/Toucan-1.5M}}} 
}}}

\newcommand{\dataname}[0]{\textsc{Toucan~}}
\newcommand{\datanamen}[0]{\textsc{Toucan}}
\definecolor{taskannocolor}{HTML}{e6d5a3}
\definecolor{trajannocolor}{HTML}{b1d8e5}

\iclrfinalcopy
\begin{document}

\maketitle

\begin{abstract}
Large Language Model (LLM) agents are rapidly emerging as powerful systems for automating tasks across domains. Yet progress in the open-source community is constrained by the lack of high quality permissively licensed tool-agentic training data. Existing datasets are often limited in diversity, realism, and complexity, particularly regarding multi-tool and multi-turn interactions. To address this gap, we introduce \datanamen, the largest publicly available tool-agentic dataset to date, containing 1.5 million trajectories synthesized from nearly 500 real-world Model Context Protocols (MCPs). Unlike prior work, \dataname leverages authentic MCP environments to generate diverse, realistic, and challenging tasks with trajectories involving real tool execution. Our pipeline first produces a broad spectrum of tool-use queries using five distinct models, applies model-based quality filtering, and then generates agentic trajectories with three teacher models using two agentic frameworks. Rigorous rule-based and model-based validation ensures high-quality outputs. We also introduce three extension mechanisms to further diversify tasks and simulate multi-turn conversations. Models fine-tuned on \dataname outperform larger closed-source counterparts on the BFCL V3 benchmark and push the Pareto frontier forward on MCP-Universe Bench.
\end{abstract}

\section{Introduction}

% \zhangchen{Hi? Can anyone proof read this version add more references?}

Large language models (LLMs) have become integral to AI applications, with LLM agents emerging as powerful systems for automating complex tasks across diverse domains \cite{li2024llm}. There is growing excitement about the potential of LLM agents to unlock new levels of automation across industries \citep{ferrag2025llmreasoningautonomousai,bousetouane2025agenticsystemsguidetransforming}. These agents handle multi-step workflows that require discovering the right tools from potentially large toolsets, calling them correctly with appropriate parameters, handle tool failures gracefully, and synthesizing results into accurate, context-aware responses \cite{xu2025llm}. Recent advancements, such as the Model Context Protocol (MCP) \citep{mcp-antrophic-blog}, have streamlined tool integration by providing standardized interfaces, enabling seamless connections between LLMs and real-world environments and simplifying the process for LLM agents to discover, invoke, and execute external tools.

Despite these advancements, progress in the open-source community is constrained by the lack of high-quality, permissively licensed \textbf{tool-agentic data} for training more capable agentic LLMs. 
An instance of tool-agentic data comprises a task-trajectory pair, where trajectories capture sequences of planning, tool calls, tool responses, and the final model response.
While previous efforts \citep{qin2023toolllmfacilitatinglargelanguage, liu2024apigenautomatedpipelinegenerating,liu2025toolacewinningpointsllm,prabhakar2025apigenmtagenticpipelinemultiturn} have introduced datasets covering various tool-calling scenarios, they suffer from several limitations: restricted tool diversity, lack of authentic tool responses, focus on single-turn conversations between users and models, or insufficient scale, all of which constrain effective training of agentic capabilities.
There is an urgent need for comprehensive, high-quality datasets that capture the full spectrum of tool-agentic interactions observed in production environments.

In this work, we bridge this gap by introducing \datanamen, the largest publicly available tool-agentic dataset to date, comprising 1.5 million trajectories synthesized from nearly 500 real-world MCP servers. Unlike prior approaches that rely on simulated or limited toolsets, \dataname leverages authentic MCP environments with more than 2,000 tools to generate diverse, realistic, and challenging tasks spanning parallel and multi-step tool calls, as well as multi-turn conversations.
Our pipeline begins by producing a broad spectrum of tool-use tasks using five distinct models with MCP server specifications, followed by model-based quality filtering to ensure relevance and difficulty. We then generate agentic trajectories with three teacher models, incorporating rigorous rule-based and model-based checks for high-quality outputs, including verification of tool execution and response accuracy. 
Our pipeline also integrates extensions to generate additional tasks targeting edge case scenarios, interactive conversations, and multi-turn dialogues.

Our experiments demonstrate the effectiveness of \dataname in enhancing LLM agentic capabilities. Models fine-tuned on \dataname surpass closed-source counterparts on the BFCL V3 benchmark \citep{patil2025bfcl}, achieving superior performance in function calling accuracy across single-turn and multi-turn scenarios. Furthermore, they show substantial improvements on $\tau$-Bench \citep{yao2024taubenchbenchmarktoolagentuserinteraction} and $\tau^2$-Bench \citep{barres2025tau2benchevaluatingconversationalagents}, with gains in tool selection, execution fidelity, and multi-turn reasoning under dynamic user interactions. On the recent MCP-Universe benchmark \citep{mcpuniverse}, which evaluates LLMs on 231 realistic tasks using 11 real-world MCP servers, \datanamen-tuned models achieve state-of-the-art performance within their parameter class, consistently outperforming leading models of comparable size.
In summary, the contributions of our work are:

\begin{itemize}[itemsep=0pt, parsep=0pt, topsep=0pt, leftmargin=1em]
  \item \textbf{\dataname Dataset}. The largest open-source tool-agent training dataset, covering parallel and multi-step tool calls, multi-turn dialogues, and edge-case tool use. Recent reports on frontier LLM development, such as Kimi-K2 \citep{kimiteam2025kimik2openagentic} and GLM-4.5 \citep{5team2025glm45agenticreasoningcoding}, highlight the value of large-scale trajectories with broad domain coverage, and \dataname provides an open-source alternative that bridges this gap.
  \item \textbf{\dataname Pipeline}. A pipeline that leverages any MCP specifications to generate diverse tool-agent trajectories, supports tool execution through MCP servers, and can be seamlessly extended to new tools via the MCP standard.
  \item \textbf{\dataname Checkpoints}. Our experiments demonstrate that models fine-tuned on \dataname mixtures surpass closed-source counterparts on the BFCL V3 and MCP-Universe benchmarks.
\end{itemize}

\section{Related Work}

\begin{table}[t]
  % \vspace{-0.1em}
  \centering
  \renewcommand{\arraystretch}{0.5}
  \caption{\dataname comparison to open-source tool-agentic datasets. Comparison comprises total trajectories, tool calling scenarios ([S]ingle, [P]arallel, [M]ulti[S]tep) including no-tool-use edge case (irrelevance[IR]), number of multi-turn conversations, and other details about data generation. Note $-$ indicates information not publicly available.}
  % \vspace{-0.5em}
  \label{tab:comparative-datasets-table}
  \resizebox{\textwidth}{!}{%
    \begin{tabular}{lccccc}
      \toprule
      \textbf{Dataset}& 
      % \textbf{Tools}& 
      % \textbf{Domains}&
      \textbf{Trajectories}&
      \makecell[l]{\textbf{Tool-Call}\\ \textbf{Scenarios}}& 
      % \makecell[l]{\textbf{Single}\\ \textbf{tool-call}}&
      % \makecell[l]{\textbf{Parallel}\\ \textbf{tool-call}}&
      % \makecell[l]{\textbf{Multi-step}\\ \textbf{tool-call}}&
      % \makecell[l]{\textbf{Single}\\ \textbf{turn}}&
      \makecell[l]{\textbf{Multi}\\ \textbf{Turn}}& 
      \textbf{Tool Specs}&
      \textbf{Tool Response} \\ 
      \midrule
      \makecell[l]{APIGent-MT-5K \citep{prabhakar2025apigenmtagenticpipelinemultiturn}}  &\num{5000}   &S P MS IR&\num{5000}  &From $\tau$-Bench &Executed \\[1em] % \hline
      \makecell[l]{ToolACE \citep{liu2025toolacewinningpointsllm}}                       &\num{11300}  &S P MS IR&\num{509}   &Synthetic         &Simulated\\[1em] % \hline
      \makecell[l]{Hermes Function-Calling V1 \citep{Hermes-Function-Calling-Dataset-V1}} &\num{11570}  &S P MS IR&\num{1890}  &Synthetic         &Executed       \\[1em] % \hline
      \makecell[l]{Nemotron (Tools) \citep{NemotronPostTrainingDatasetV1}}               &\num{310051} &S P MS  --&\num{199610}&--                &--       \\[1em]% \hline
      \dataname (This Work)                                                                      &\num{1527259}&S P MS IR&\num{567262}&Real              &Executed \\ \bottomrule
    \end{tabular}%
    \vspace{-1em}
  }
\end{table}

\textbf{The past: Tool-calling datasets and benchmarks for LLMs.}
Early tool-calling datasets enabled LLMs to interact with tools like REST APIs and ML functions. The Gorilla project \citep{patil2023gorillalargelanguagemodel} demonstrated that fine-tuning on such data enhances tool-use over vanilla models, introducing the BFCL benchmark \citep{patil2025bfcl} as a standard. ToolAlpaca \citep{tang_toolalpaca_2023} offered cost-effective synthetic data with lower quality, while ToolLLM \citep{qin2023toolllmfacilitatinglargelanguage} expanded to 16,000+ APIs across domains. API Pack \citep{guo2025apipackmassivemultiprogramming} added cross-language diversity (Python, Java, C++), and API Blend \citep{basu2024apiblendcomprehensivecorporatraining} optimized dataset mixtures for robustness, laying the foundation for tool-agent advancements. \textcolor{black}{More recently, APIGen has focused on domain diversification, contributing a training dataset covering 21 domains \cite{liu2024apigenautomatedpipelinegenerating}.}

\textbf{The present: Tool-calling benchmarks and datasets for LLM-agents.}
Recent research has shifted toward training LLM agents for effective tool use, exemplified by models like Kimi-K2 \citep{kimiteam2025kimik2openagentic} and GLM-4.5 \citep{5team2025glm45agenticreasoningcoding}, with performance assessed via benchmarks such as BFCL \citep{patil2025bfcl}, $\tau$-Bench \citep{yao2024taubenchbenchmarktoolagentuserinteraction}, and ACEBench \citep{chen_acebench_2025}. BFCL covers diverse scenarios including parallel, multi-step, and multi-turn tool use, while $\tau$-Bench focuses on realistic user-agent-tool interactions. ACEBench enhances evaluation by addressing edge cases and including a subset for tool-agent trajectories. Despite these advances, open-source training data for tool-agent trajectories remains limited. \textcolor{black}{Existing datasets \citep{Hermes-Function-Calling-Dataset-V1,liu2025toolacewinningpointsllm,prabhakar2025apigenmtagenticpipelinemultiturn,NemotronPostTrainingDatasetV1} either lack dataset curation transparency, are small in size for SFT, or simulate tool responses via LLMs}. Table~\ref{tab:comparative-datasets-table} compares existing tool-agentic datasets with \datanamen, which, at 1.5 million trajectories, offers the largest dataset, featuring extensive multi-turn dialogues, all tool-use scenarios, critical edge cases, and authentic tool responses from real-world environments.

\newcommand{\cmark}{\textcolor{green}{\ding{51}}}%
\newcommand{\xmark}{\textcolor{red}{\ding{55}}}%

\textbf{The future: MCP benchmarks and datasets.} As concurrent work, recent MCP benchmarks \citep{gao2025mcpradarmultidimensionalbenchmarkevaluating,wang2025mcpbenchbenchmarkingtoolusingllm,mcpuniverse,mcpmark_2025,guo2025mcpagentbenchevaluatingrealworldlanguage,yin2025livemcp,liu2025mcpevalautomaticmcpbaseddeep,yan2025mcpworldunifiedbenchmarkingtestbed,scale2025mcp} aim to rigorously assess LLMs in tool-use settings beyond simple correctness. For instance, MCP-Radar \citep{gao2025mcpradarmultidimensionalbenchmarkevaluating} employs a five-dimensional evaluation including accuracy, tool selection efficiency, resource usage, parameter construction, and execution speed across software engineering, math, and problem-solving tasks with 300 queries and 42 MCP servers. Similarly, MCP-Bench \citep{wang2025mcpbenchbenchmarkingtoolusingllm} evaluates multi-step reasoning over 28 MCP servers and 250 tools, while MCP-Universe \citep{mcpuniverse} focuses on execution-based metrics in six real-world domains. These advancements underscore the need for comprehensive training datasets to support the development of robust, open-source LLM agents.

\section{\datanamen: Scaling Tool-Agentic Data with Real World MCPs}
\label{sec:toucan-overview}

\begin{wrapfigure}{r}{0.37\textwidth}
  \vspace{-4.5em}
  \centering
  \includegraphics[width=\linewidth]{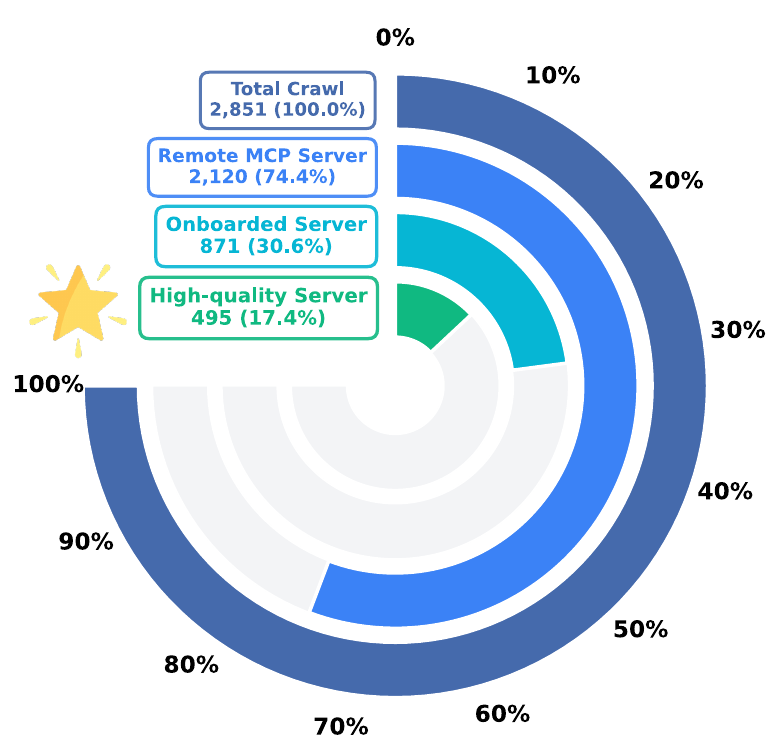}
        \caption{MCP servers filtering process}
        \label{fig:server-num}
  \vspace{-2.5em}
\end{wrapfigure}

\subsection{\dataname Generation Pipeline}
\begin{figure}[t!]
    \centering
    % \vspace{-1em}
    \includegraphics[width=1.0\linewidth]{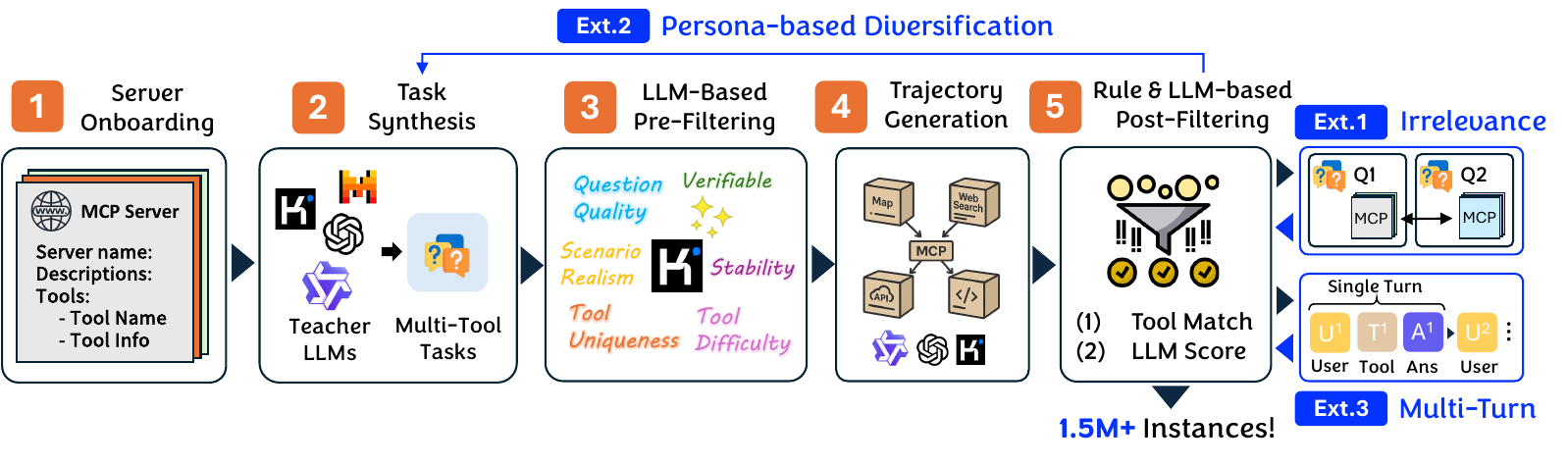}
    \vspace{-1em}
    \caption{The \dataname construction pipeline: A systematic five-stage process from MCP server onboarding through trajectory filtering, with three extensions for enhancing data diversity and realism.}
    \label{fig:toucan-mcp-pipeline}
    \vspace{-1em}
\end{figure}

\dataname is a comprehensive dataset comprising over 1.5 million tool-agent trajectories constructed using real-world tools from MCP servers. Each instance in our dataset contains a task description, a complete agent trajectory with its associated tools, quality and classification annotations, as well as comprehensive metadata. Appendix~\ref{app:dataset-schema} provides a detailed schema description and demonstration samples.
The construction of \dataname follows a systematic five-stage pipeline: MCP server onboarding, task synthesis, task filtering, trajectory generation, and trajectory filtering. Additionally, we implement three extension mechanisms to further enhance data diversity and realism. Figure~\ref{fig:toucan-mcp-pipeline} illustrates the complete construction pipeline. We detail each stage below.

% Our pipeline first produces a broad spectrum of tool-use tasks using five distinct models, applies model-based quality filtering, then generates agentic trajectories with three teacher models. Rigorous rule and model-based checks ensure high-quality outputs.

\begin{wrapfigure}{r}{0.43\textwidth}
  \vspace{-1.8em}
  \centering
  \includegraphics[width=1\linewidth]{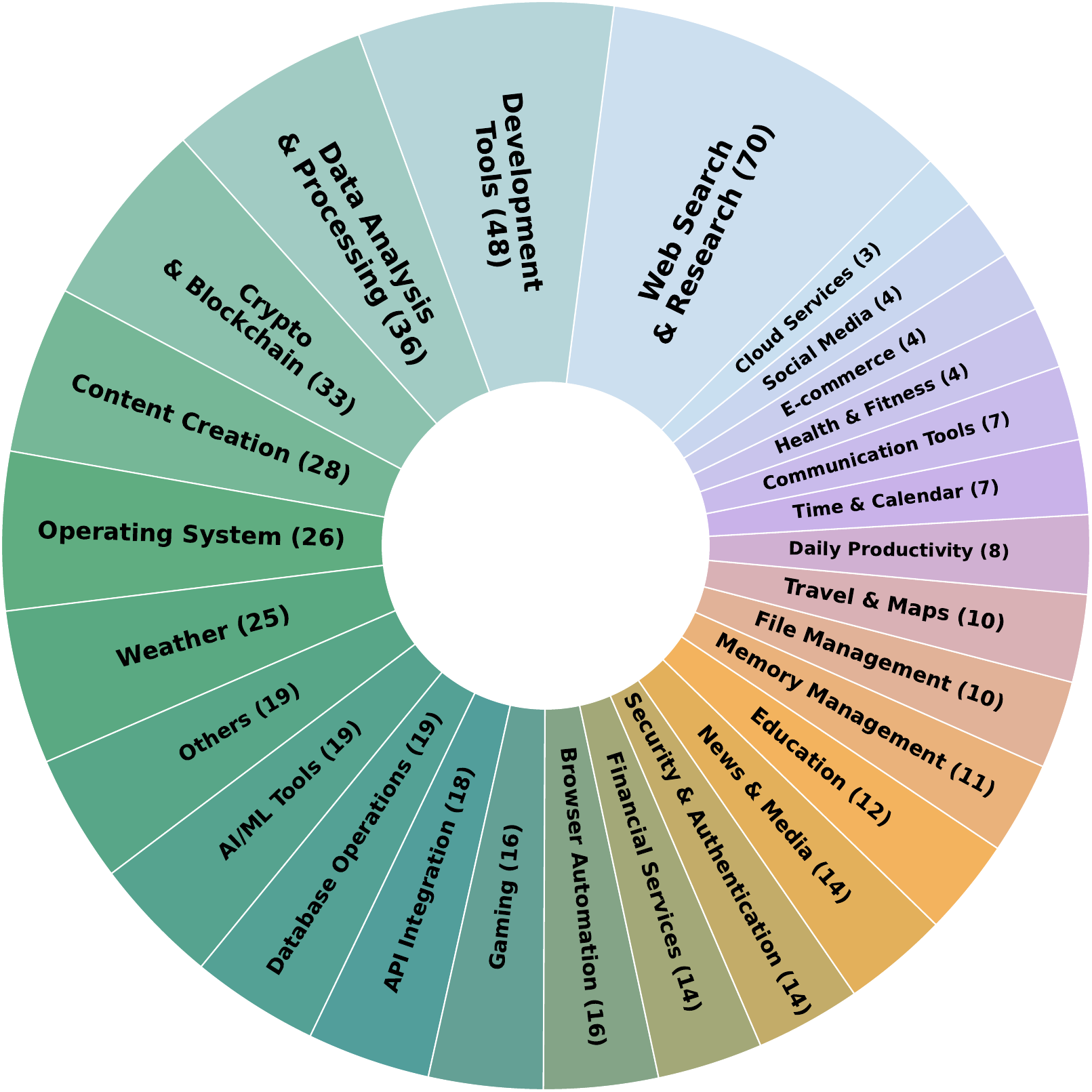}
  \vspace{-0.5em}
  \caption{MCP servers distribution by domain, covering a wide range of categories. Values in parentheses indicate the number of servers belonging to each category.}
  \label{fig:mcp-servers-category}
  \vspace{-2em}
\end{wrapfigure}

\textbf{\textcolor{orange!90!black}{Stage 1: MCP Server Onboarding.}}
%  \begin{itemize}
%     \item Crawling
%     \item Filtering: rule-based (connectivity and stability check, onboarding without special configurations)
%     \item MCP Server category annotation
% \end{itemize}
To generate questions from diverse environments, the initial step involves onboarding as many high-quality MCP servers as possible. We sourced MCP server specification files from GitHub and Smithery \footnote{https://smithery.ai/}, a platform and registry for MCP servers that encapsulate modular execution environments. Each MCP server is accompanied by a structured JSON document detailing metadata about the server with a machine-readable definition of the tools it provides. From an initial crawl yielding approximately 2,800 MCP servers, we applied two key filtering criteria: (1) retaining only remote MCP servers accessible via streamable HTTP to ensure compatibility with trajectory generation, and (2) excluding servers requiring third-party credentials (e.g., API keys) for tool invocation to maintain accessibility and reproducibility. This process reduced the dataset to 30.6\% (871 servers). As a final step, we generated a small subset of test questions to evaluate each tool within the MCP servers, subsequently filtering out servers with problematic tools that returned error messages or failed to function correctly. This rigorous curation process resulted in a refined set of 495 high-quality MCP servers spanning diverse domains and functionalities.
Figure~\ref{fig:server-num} depicts the number of MCP servers retained at each filtering stage. Figure \ref{fig:mcp-servers-category} demonstrates the domain distribution of the final server collection across diverse categories. The domain distribution is annotated by LLMs, where prompts can be found in Appendix \ref{app:server-category-annotation-prompt}.

% \begin{figure}[t!]
% \vspace{-2em}
%     \centering
%     \begin{subfigure}[t]{0.34\textwidth}
%         \includegraphics[width=\linewidth]{images/server_num.pdf}
%         \caption{MCP servers filtering process}
%         \label{fig:server-num}
%     \end{subfigure}
%     \hfill
%     \begin{subfigure}[t]{0.33\textwidth}
%         \includegraphics[width=1.0\linewidth]{images/mcp_primary_labels_ring_internal.pdf}
%         \caption{MCP servers distribution by domain}
%         \label{fig:mcp-servers-category}
%     \end{subfigure}
%     % \caption{MCP servers analysis: (a) summarizes MCP servers filtering process, (b) MCP servers distribution by domain, (c) tools number distribution per MCP server.}
%     \caption{Overall statistics of \dataname dataset.}
%     \vspace{-2em}
%     \label{fig:mcp-servers-analysis}
% \end{figure}

% Tool target selection strategies
% Other strategies: persona
% Teacher models

%  \begin{itemize}
%     \item Single server:  single tool, multi-tool; balance distribution according to server usage)
%     \item Multi Server (same category, different category, default as multi-tool)
%     \item Featured Server (a list of aprox. 20 high quality and representative servers, send to LLMs and let it choose whatever makes sense， default as multi-tool)
% \end{itemize}

\textbf{\textcolor{orange!90!black}{Stage 2: Task Synthesis.}} The next step involves synthesizing high-quality tasks from MCP servers, where each task comprises a question and the desired tool names from the MCP servers. The key challenge is ensuring that tasks are challenging, realistic, and cover edge cases. Therefore, we design diverse sampling strategies based on MCP server usage number from Smithery and server functionalities. To avoid potential bias from individual models, we utilized five open-source LLMs (\texttt{Mistral-Small}, \texttt{DevStral-Small}, \texttt{GPT-OSS}, \texttt{Kimi-K2}, and \texttt{Qwen3-32B}) as task generators to construct synthetic tasks (see the prompts in Appendix~\ref{app:task-generation-prompt-for-single-server}). We apply the following three strategies to synthesize tasks, where the maximum number of tools is set to $N=3$ in our experiments:

\textbf{Single Server:} For a given MCP server, we synthesize tasks requiring the use of 1 to $N$ tools, ensuring a balanced selection distribution guided by server usage statistics to reflect real-world applicability.

\textbf{Multi-Server:} Leveraging LLM-based domain annotations derived from MCP metadata, we first sample $N$ MCP servers from either the same or different categories. We then prompt LLMs to conduct a server analysis, outlining potential workflows that integrate tools across these servers, targeting two to $N$ specific tools, and subsequently generating tasks that leverage functionalities from multiple servers.

\textbf{Featured Server:} Based on the original MCP file metadata, we manually selected 25 representative MCP servers from various domains, with the complete list available in Appendix~\ref{app:featured-server-list}. In this approach, we provide all MCP server metadata within the context, specify an expected number of tools, and allow the LLM to freely explore combinations, devise realistic scenarios, select the necessary tools, and create comprehensive tasks.

\textbf{\textcolor{orange!90!black}{Stage 3: Task Filtering.}} 
% LLM-based annotation: quality, realisim, stability, uniqueness, difficulty
To ensure the quality of synthesized tasks, this stage involves annotating tasks across six dimensions and filter out suboptimal instances. We employed the \texttt{Kimi-K2} model as the annotator, which was selected for its optimal balance between correlation with human annotations and cost efficiency. The correlation statistics are detailed in Appendix~\ref{app:llm-annotation}, and the prompt template is provided in Appendix~\ref{app:task-annotation-prompt-for-single-server}. Each dimension is rated on a 1-5 Likert scale. The detailed evaluation metrics are as follows:

\begin{itemize}[itemsep=0pt, parsep=0pt, topsep=0pt, leftmargin=1em]
\item \textit{Tool Selection Difficulty:} Judges the difficulty of selecting the required tools from provided tools.
\item \textit{Tool Selection Uniqueness:} Assesses the uniqueness of the selected tool combination relative to the available tools, and whether viable alternatives could also solve the task.
\item \textit{Question Quality:} The task's overall quality, reflected by its clarity, specificity, and effectiveness.
\item \textit{Scenario Realism:} Evaluates the authenticity and realism of the task scenario.
\item \textit{Verifiable:} Evaluates how easily the final model answer can be verified given the question.
\item \textit{Stability:} Evaluates whether tool outputs remain consistent over time, across geolocation, and under stochastic variation.
\end{itemize}

\textbf{\textcolor{orange!90!black}{Stage 4: Trajectory Generation.}}
This step involves collecting trajectories including tool calls, tool responses, and reasoning steps in agentic environments given tasks synthesized and filtered from the previous steps. To ensure diversity, we employed three LLMs from different families (\texttt{GPT-OSS-120B}, \texttt{Kimi-K2}, and \texttt{Qwen3-32B}) in combination with two agent frameworks (\texttt{Qwen-agent} and \texttt{OpenAI-agent}) to produce high-quality agentic trajectories. The models are deployed remotely and accessed by the agent frameworks via streamable HTTP.

\textbf{\textcolor{orange!90!black}{Stage 5: Rule\&LLM-Based Post-Filtering.}}
The trajectory filtering process combines rule-based verifiers with LLM-driven annotations to ensure high quality. Rule-based heuristics exclude trajectories that fail to start the agent or connect successfully with remote MCP servers, do not contain tool calls, exhibit failures in tool responses, or contain local file system paths. We also validate whether the trajectory uses the required tools specified by the task in the correct sequence, and report both the \textit{desired tool use percentage} (coverage of required tools) and \textit{order correctness} (adherence to expected sequence) metrics. We then employ \texttt{GPT-OSS-120B} as a judge to annotate each trajectory in terms of completeness and conciseness. The annotation prompt is provided in Appendix~\ref{app:trajectory-annotation-prompt-for-single-server}, with metric definitions as follows:
\begin{itemize}[itemsep=0pt, parsep=0pt, topsep=0pt, leftmargin=1em]
\item \textit{Completeness:} Judges whether the assistant fulfills the user's request end-to-end.
\item \textit{Conciseness:} Judges whether the task is solved with the minimum necessary steps and verbosity.
\end{itemize}
This dual-stage filtering approach ensures that only high-quality, concise, and executable trajectories are retained in the final dataset.

\subsection{\dataname Extensions}\label{sec:toucan-extensions}

While the core pipeline generates high-quality trajectories, these are single-turn interactions between user and agent without follow-ups, which limits their practical applicability to real-world scenarios. In addition, since all available tools are contextually relevant, tool selection becomes trivial for LLMs, resulting in relatively low difficulty. To address these limitations and enhance the dataset's versatility, we apply three distinct procedures post-core pipeline (Steps 1 to 5) to generate new instances targeting specific objectives.

\textbf{\textcolor{blue!60!black}{Ext.1: Irrelevance.}} To reduce hallucination, it is critical to train models to reject unanswerable queries or seek alternative solutions when desired tools are unavailable. To achieve this, we systematically generate queries unsolvable with the current toolset (Ext1 in Figure~\ref{fig:toucan-mcp-pipeline}) by shuffling MCP server metadata across instances and repeating the task generation step.

\textbf{\textcolor{blue!60!black}{Ext.2: Persona-based Diversification.}} We implement persona-based diversification (Ext2 in Figure~\ref{fig:toucan-mcp-pipeline}) to create varied task versions. This involves two strategies: one enhances diversity by introducing new contexts and personas, while the other increases task complexity through additional constraints, all while utilizing the same target tools. This diversification process produces tasks similar yet distinct from those in the core pipeline. The prompts are detailed in Appendix~\ref{app:task-diversification-prompts}.

\textbf{\textcolor{blue!60!black}{Ext.3: Multi-Turn.}} Recognizing that real-world user-agent-tool interactions seldom conform to single-turn conversations \cite{yao2024taubenchbenchmarktoolagentuserinteraction}, we introduce a self-simulation pipeline to generate multi-turn dialogues using the trajectory generation model. This is achieved through two methods: (1) splitting complex tasks requiring multi-tool coordination into sequential sub-questions, and (2) extending existing conversations by providing LLMs with context to formulate follow-up queries.

Finally, we repeat the core pipeline from steps~2 to 5 to build full trajectories with the new tasks. In the case of irrelevant tasks (Ext.1), we tighten trajectory filters to retain only instances with zero tool calls. Together, these data extensions yield a more realistic and robust \dataname dataset that covers all relevant tool-use scenarios and user question styles.

\begin{figure}[t!]
% \vspace{-2em}
    \centering
    \includegraphics[width=1.0\linewidth]{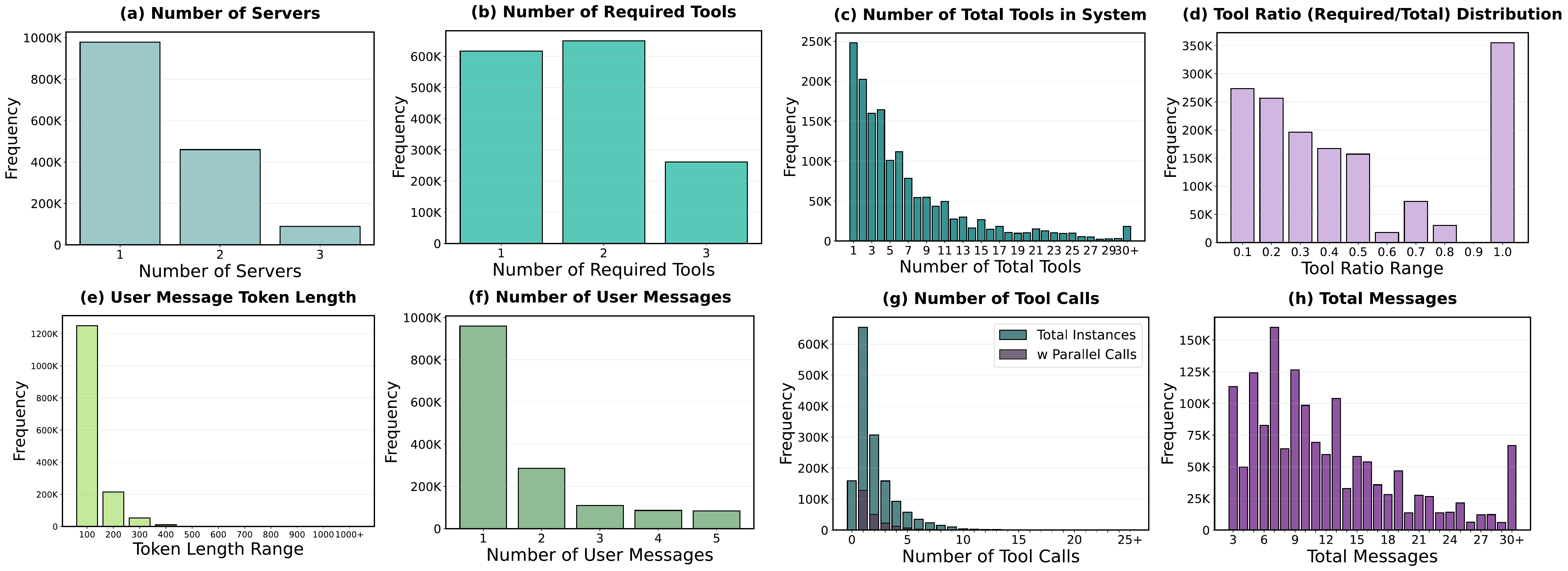}
    \caption{The figures above illustrate the \dataname dataset analysis. Subfigure (a) and (b) provide statistics on the number of servers and required tools per instance, highlighting \dataname's comprehensive coverage of multi-server and multi-tool tasks. Subfigures (c) and (d) reveal that most tasks include more tools in the context than the targeted tools, underscoring the non-trivial tool selection challenges. Subfigure (e) displays the length of user messages in tokens. Subfigures (f) and (h) demonstrate the multi-turn nature of the tasks, characterized by extended and diverse interactions among users, agents, and tools. Subfigure (g) demonstrates that \dataname encompasses both single and parallel tool calls, which enhance the dataset's versatility in capturing diverse agent-tool interaction patterns.}
    \label{fig:toucan-dataset-analysis}
    \vspace{-1em}
\end{figure}

\begin{figure}[h]
    \centering
    \begin{minipage}[t]{0.41\textwidth}
        \centering
        \includegraphics[width=\linewidth]{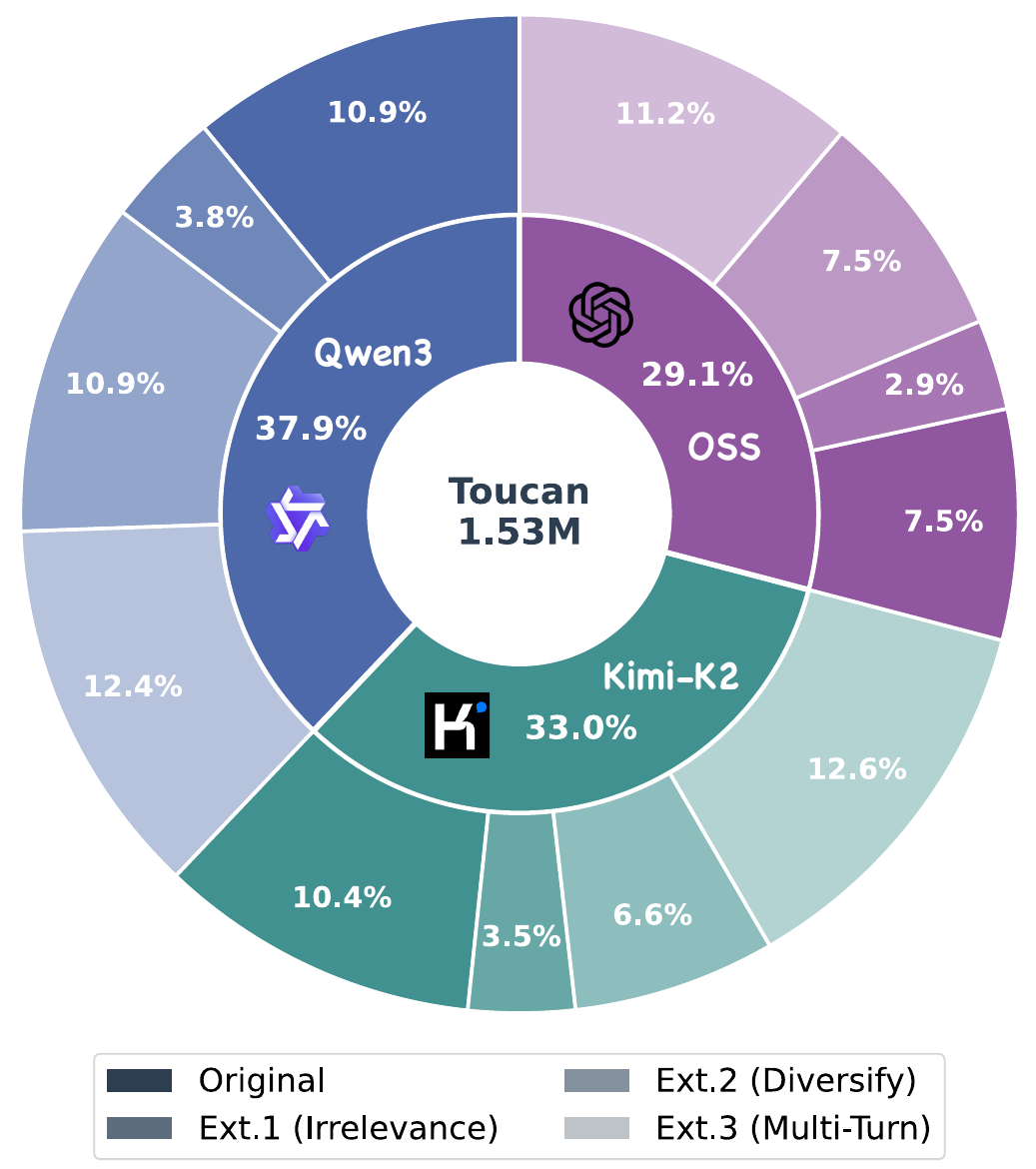}
        \caption{\dataname Subset Statistics}
        \label{fig:toucan-subset-stats}
    \end{minipage}
    \hfill
    \begin{minipage}[t]{0.58\textwidth}
        \centering
        \includegraphics[width=\linewidth]{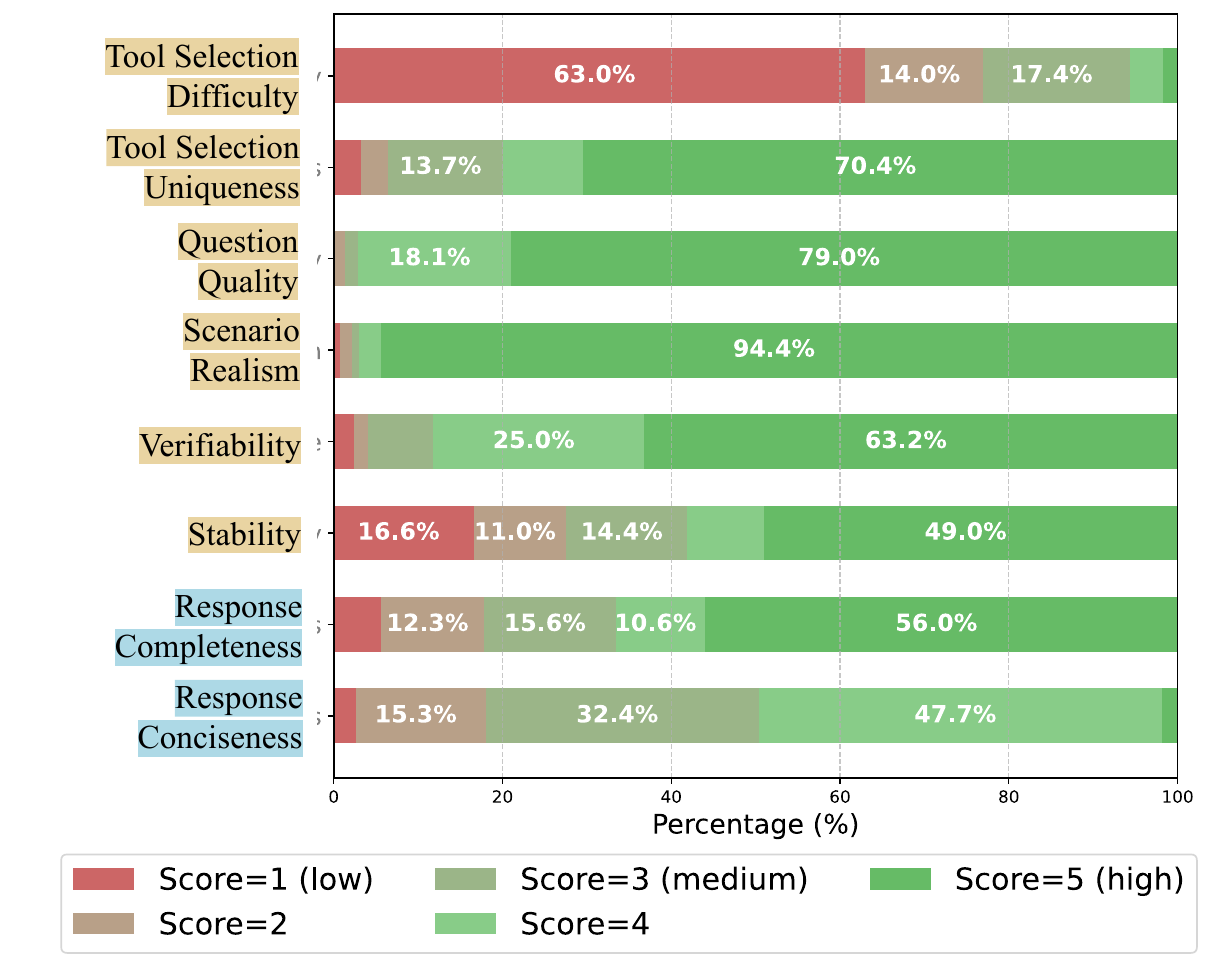}
        \caption{\dataname Quality Statistics}
        \label{fig:toucan-quality-analysis}
    \end{minipage}
    \vspace{-0.5em}
\end{figure}

\subsection{Data Analysis}

This section analyzes the generated \dataname dataset from statistical analysis and LLM-based quality assessment.

\textbf{Statistical Analysis of \dataname.} We conduct comprehensive statistical analysis of MCP servers and data instances. The top MCP servers used in \dataname and tool statistics within each MCP servers are presented in Appendix~\ref{app:top-mcps}. Figure~\ref{fig:toucan-dataset-analysis} provides a comprehensive analysis of the \dataname dataset. We observe that \dataname provides comprehensive coverage of multi-server and multi-tool tasks, and includes multi-turn conversations among users, agents, and tools. Additionally, most tasks contain more tools in the context than the required target tools, indicating non-trivial tool selection requirements. 
Figure~\ref{fig:toucan-subset-stats} presents the subset statistics of \dataname across different trajectory generator LLMs and data partitions.
We also provide embedding visualization of \dataname using UMAP projection in Appendix~\ref{app:embedding-visualization}, demonstrating the wide domain coverage of \datanamen.

\textbf{Quality Assessment of \datanamen.} Figure \ref{fig:toucan-quality-analysis} presents a statistical analysis conducted by an LLM-as-a-judge on \datanamen. From the task perspective (labels in \raisebox{3pt}{\colorbox{taskannocolor}{}}), we observe that the majority of tasks exhibit exceptionally high question quality and scenario realism, indicating robust task design and alignment with real-world applications. Additionally, the dataset features a mixed difficulty range, encompassing both simple and challenging tasks. From the response perspective (label in \raisebox{3pt}{\colorbox{trajannocolor}{}}), we find that trajectory quality is satisfactory, with most scores at or above 3 (medium) across both completeness and conciseness metrics.

\section{Experiments}

In this section, we demonstrate the performance of \dataname by performing supervised fine-tuning (SFT) on baseline models of different sizes. We then compare the fine-tuned models' performance against existing model baselines across several widely used agentic tool-call benchmarks.

\subsection{Experiment Setup}

\textbf{Model and Baseline Setup.} We perform supervised fine-tuning on \texttt{Qwen2.5-7B-Instruct}, \texttt{Qwen2.5-14B-Instruct}, and \texttt{Qwen2.5-32B-Instruct} \citep{qwen2.5} to demonstrate the efficacy of \dataname across models of varying sizes. Detailed fine-tuning parameters are provided in Appendix \ref{app:fine-tune-parameter}. We benchmark the performance of our fine-tuned models against models of comparable or larger scales, including \texttt{DeepSeek-V3} \cite{deepseekai2025deepseekv3technicalreport}, \texttt{Qwen2.5-72B-Instruct}, \texttt{Qwen3-235B-A22B}, \texttt{Qwen3-32B} \cite{yang2025qwen3technicalreport}, and closed-source OpenAI models such as o3-mini, GPT-4.1, and GPT-4.5-Preview.

\begin{table}[t!]
  \vspace{-1em}
  \centering
  \caption{This table compares the performance of \datanamen-tuned models and baselines on the BFCL-V3 benchmark. We observe that \dataname remarkably improves baseline model performance through supervised fine-tuning (SFT) and enables smaller models to outperform larger models across different evaluation aspects.}
  % \vspace{-1.5em}
  \label{tab:bfclv3-results-table}
  \resizebox{\textwidth}{!}{%
\begin{tabular}{@{}lcccccc@{}}
\toprule
\textbf{Model}                          & \textbf{Overall} & \multicolumn{2}{c}{\textbf{Single Turn}} & \textbf{Multi Turn} & \multicolumn{2}{c}{\textbf{Hallucination}} \\
                                       \cmidrule(lr){2-2} \cmidrule(lr){3-4} \cmidrule(lr){5-5} \cmidrule(lr){6-7}
                                       &                  & \textit{Non-live (AST)}        & \textit{Live (AST)}       &                     & \textit{Relevance}           & \textit{Irrelevance}         \\
\midrule  
DeepSeek-V3                            & 64.71\%          & 88.54\%               & 77.34\%          & 29.87\%             & \textbf{83.33\%}             & 76.49\%              \\
Qwen2.5-72B-Instruct              & 64.37\%          & 87.56\%               & 78.68\%          & 29.38\%             & 72.22\%             & 77.41\%              \\
Qwen3-235B-A22B                        & 67.94\%          & 87.90\%               & 77.03\%          & 40.12\%             & \textbf{83.33\% }            & 76.32\%              \\
Qwen3-32B                              & 69.25\%          & \textbf{88.90\%}               & 77.83\%          & 43.12\%             & 72.22\%             & 75.79\%  \\
% ToolACE-2-8B &  68.73\% & 87.58\%   & 80.05\% & 37.00\% & 72.22\% & 90.11\% \\
% xLAM-2-70b-fc-r (FC)                   & 78.45\%          & 88.44\%               & 72.95\%          & 75.00\%           & 66.67\%               & 78.91\%\\
o3-Mini                                & 64.61\%          & 86.15\%               & 79.08\%          & 28.75\%             & 72.22\%             & 82.96\%              \\
GPT-4.1                                & 68.69\%          & 85.42\%               & \textbf{79.92\%}          & 40.50\%             & 77.78\%             & \textbf{85.95\%}              \\
GPT-4.5-Preview                        & 70.32\%          & 86.12\%               & 79.34\%          & 45.38\%             & 66.67\%             & 83.64\%              \\
\midrule
\midrule
%\cmidrule{1-1} \cmidrule(lr){2-2} \cmidrule(lr){3-4} \cmidrule(lr){5-5} \cmidrule{6-7}
Qwen2.5-7B-Instruct               & 55.10\%          & 84.19\%               & 72.32\%          & 12.88\%             & 72.22\%             & 67.93\%              \\
\quad\textit{with \dataname}  & \withsup{58.26\%}{3.16} & 78.52\%               & 74.50\%          & 22.62\%             & 66.67\%             & 75.18\%              \\
\midrule
Qwen2.5-14B-Instruct              & 57.69\%          & 83.38\%               & 73.70\%          & 19.75\%             & \textbf{83.33\%}             & 68.46\%              \\
\quad\textit{with \dataname} & \withsup{65.09\%}{7.40} & 85.42\%               & 76.01\%          & 35.25\%             & 72.22\%             & 75.96\%              \\
\midrule
Qwen2.5-32B-Instruct              & 61.73\%          & 85.58\%               & 76.01\%          & 26.38\%             & 72.22\%             & 72.68\%              \\
\quad\textit{with \dataname} & \withsup{\textbf{70.45\%}}{8.72} & 87.12\%               & 78.90\%          & \textbf{46.50\%}             & 77.78\%             & 78.10\%              \\
\bottomrule
\end{tabular}
\vspace{-3em}
}
\end{table}

\textbf{\dataname Setup.} Given the large volume of the full dataset, we adopted a strategy similar to \cite{Xu2025KodCodeAD} by sampling from a high-quality subset of \datanamen. This subset was selected based on the following criteria: question quality and scenario realism scores of 5, response completeness and conciseness scores of at least 4, and desired tool use percentage of 1.0 (indicating that trajectories fully utilize all required tools from the task). We performed necessary data re-balancing to ensure the dataset remains representative across different categories. The resulting SFT dataset comprises 28.3K instances from the original pipeline, 40K instances from Ext.1 (Irrelevance), 15.8K instances from Ext.2 (Diversify), and 35.2K instances from Ext.3 (Multi-Turn), totaling 119.3K instances.

\textbf{Benchmarks.} We assess the performance of \dataname across several key tool-agentic benchmarks, including BFCL V3 \cite{patil2025bfcl}, $\tau$-Bench \cite{yao2024taubenchbenchmarktoolagentuserinteraction}, $\tau^2$-Bench \citep{barres2025tau2benchevaluatingconversationalagents}, and MCP-Universe \cite{mcpuniverse}. All evaluations are conducted on an 8 $\times$ H100 server. For BFCL-V3, we use the official evaluation setup. For $\tau$-Bench and $\tau^2$-Bench, we employ \texttt{GPT-4o} as user simulators. For MCP-Universe, we configure the local evaluation environment as specified in the benchmark documentation.

\subsection{Experimental Results}

\begin{table}[t]
  \centering
  \vspace{-1.5em}
  \caption{This table presents $\tau$-Bench and $\tau^2$-Bench results for models fine-tuned on \dataname compared to their respective baselines. Improvements are observed across most evaluation scenarios.}
  \label{tab:tau-results-table}
  \resizebox{1.0\textwidth}{!}{%
\begin{tabular}{@{}lccccccc@{}}
\toprule
\textbf{Model}            & \multicolumn{3}{c}{\textbf{$\tau$-bench}}                                                                      & \multicolumn{4}{c}{\textbf{$\tau^2$-bench}}                                                                                                           \\ \cmidrule(lr){2-4} \cmidrule(lr){5-8}
                          & \multicolumn{1}{c}{\textit{Avg.}} & \multicolumn{1}{c}{\textit{Airline}} & \multicolumn{1}{c}{\textit{Retail}} & \multicolumn{1}{c}{\textit{Avg.}} & \multicolumn{1}{c}{\textit{Airline}} & \multicolumn{1}{c}{\textit{Retail}} & \multicolumn{1}{c}{\textit{Telecom}} \\\midrule
                          
Qwen2.5-7B-Instruct  & 15.03\%                           & 8.75\%                               & 21.30\%                             & 16.08\%                           & 14.00\%                              & 17.54\%                             & 16.70\%                              \\
\quad\textit{with \dataname}      & \withsup{22.48\%}{7.45}                           & 15.50\%                              & 29.46\%                             & \withsup{17.77\%}{1.69}                           & 20.00\%                              & 22.80\%                             & 10.50\%                              \\
\midrule
Qwen2.5-14B-Instruct & 30.85\%                           & 17.25\%                              & 44.46\%                             & 24.46\%                           & 12.00\%                              & 41.20\%                             & 20.18\%                              \\
\quad\textit{with \dataname}      & \withsup{35.24\%}{4.39}                           & 22.00\%                              & 48.48\%                             & \withsup{30.43\%}{5.97}                           & 22.00\%                              & 49.10\%                             & 20.18\%                              \\
\midrule
Qwen2.5-32B-Instruct  & 38.76\%                           & 26.00\%                              & 51.52\%                             & 29.40\%                           & 18.00\%                              & 49.10\%                             & 21.11\%                              \\
\quad\textit{with \dataname}      & \withsup{42.33\%}{3.57}                           & 29.00\%                              & 55.65\%                             & \withsup{31.60\%}{2.20}                           & 22.00\%                              & 52.60\%                             & 20.20\%                              \\ \bottomrule
\end{tabular}
}
\end{table}

\begin{figure}[!t]
    \centering
    \vspace{-0.5em}
    \includegraphics[width=1\linewidth]{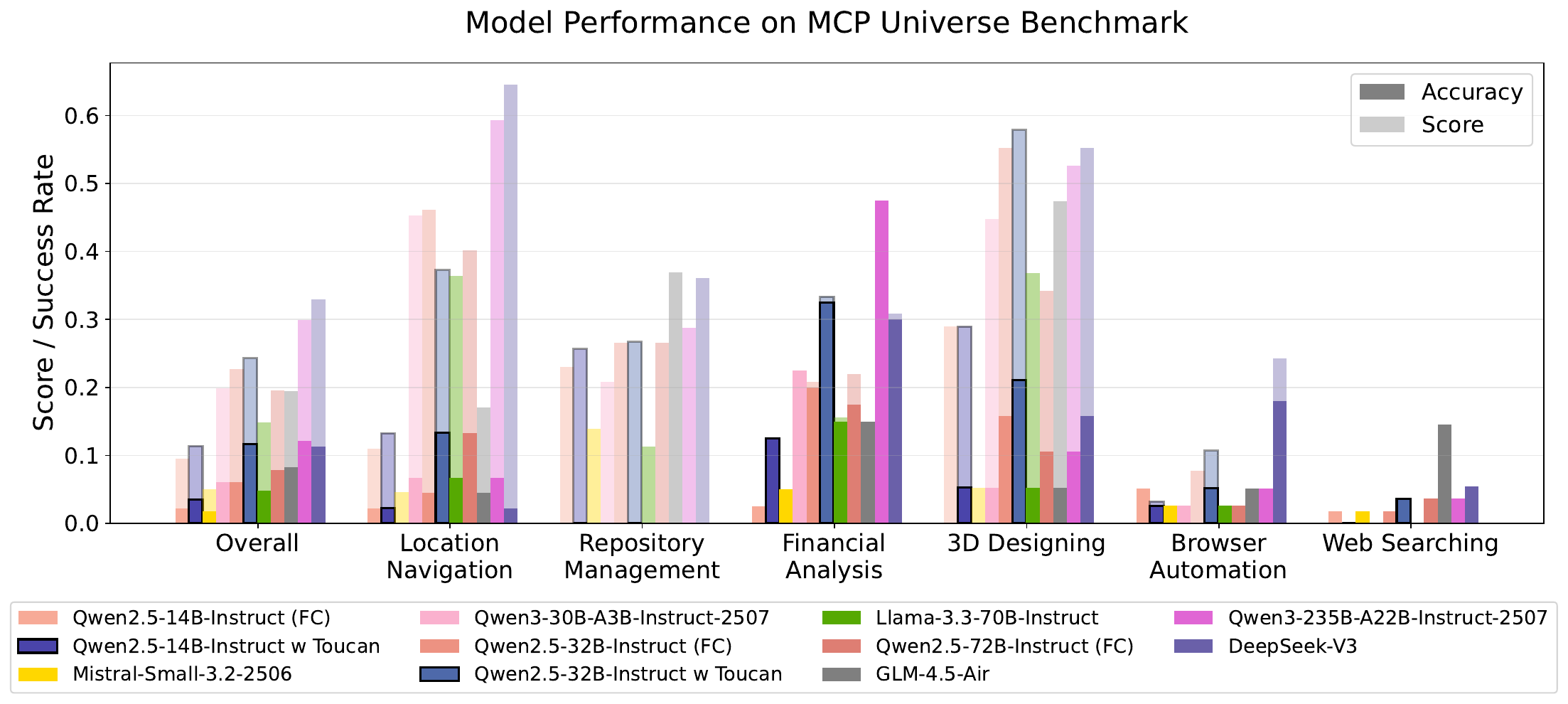}
    \caption{This figure compares the performance of \datanamen-tuned models with other open-source models on MCP-Universe \citep{mcpuniverse}. Model sizes increase from left to right. Bars with darker colors represent task success rate (full task completion), while lighter colors represent average evaluation scores considering partial task completion. \datanamen-tuned models are shown with black borders. \datanamen-tuned models outperform other models of similar sizes across most tasks.}
    \vspace{-1em}
    \label{fig:mcp-universe}
\end{figure}

\begin{wrapfigure}{r}{0.38\textwidth}
  \vspace{-3em}
  \centering
  \includegraphics[width=1\linewidth]{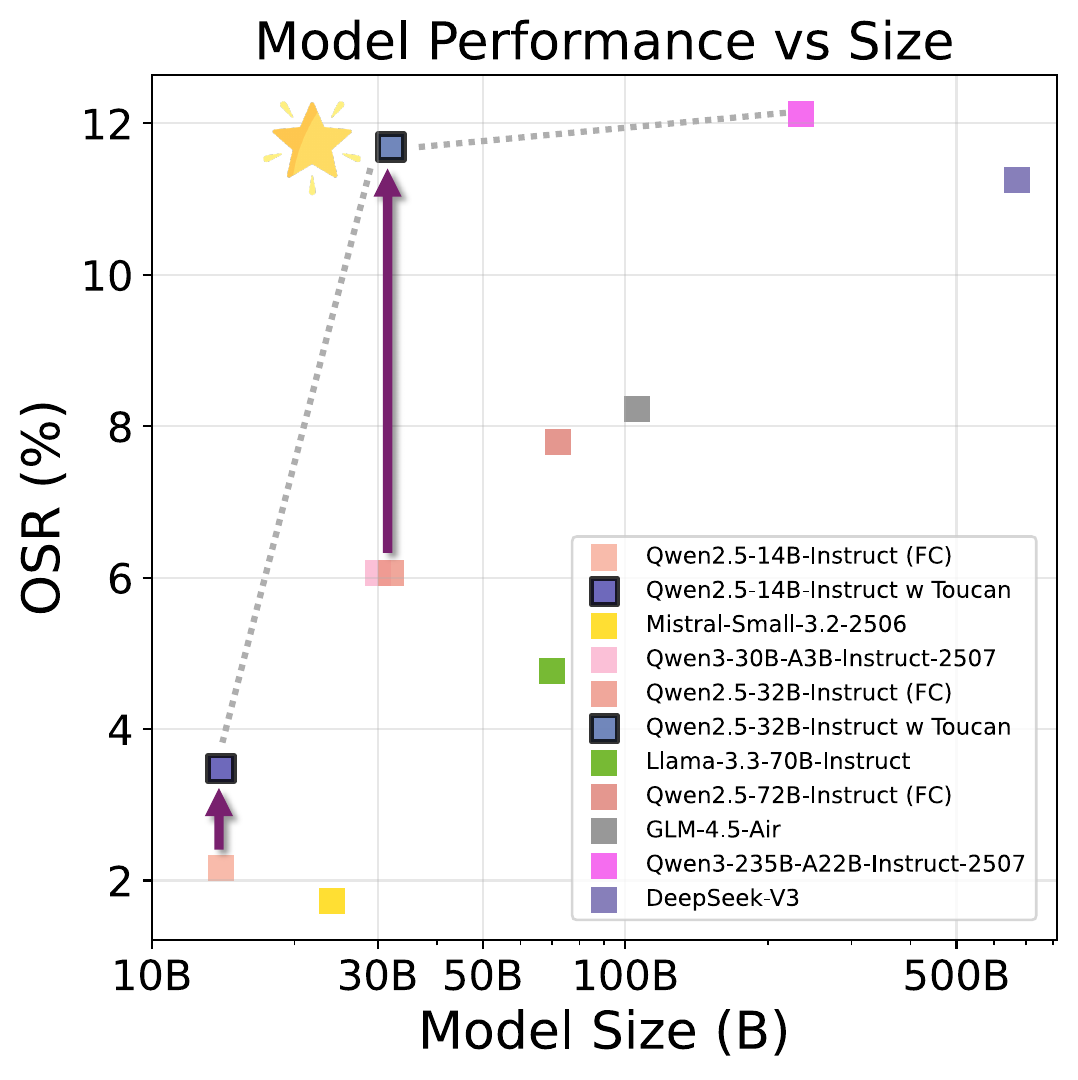}
  \vspace{-1.5em}
  \caption{Model Performance vs Size on MCP-Universe Benchmark. We report overall task success rate (OSR). Our models push the Pareto frontier forward, achieving higher OSR at smaller model sizes.}
  \label{fig:mcp-universe-pareto}
  \vspace{-2em}
\end{wrapfigure}

\textbf{\dataname Effectively Increases Agentic Tool-Calling Performance.} 
Tables~\ref{tab:bfclv3-results-table} and \ref{tab:tau-results-table} present the experimental results of models fine-tuned on \dataname across BFCL V3, $\tau$-Bench, and $\tau^2$-Bench, respectively. We make the following key observations: 
First, models fine-tuned with \dataname show performance improvements compared to baseline models without fine-tuning across almost all aspects of these three benchmarks, indicating that \dataname effectively enhances the agentic and tool-calling capabilities of models. 
Second, on BFCL V3, models fine-tuned on \dataname outperform larger production LLMs including \texttt{DeepSeek-V3} and \texttt{GPT-4.5-Preview} in average scores and achieve top performance in the \textit{multi-turn} subset. This demonstrates the effectiveness of \dataname and validates our dataset design.

\textbf{\dataname Enhances Models' Performance on Using Real-World MCP Servers.} Figure \ref{fig:mcp-universe} demonstrates a performance comparison between \datanamen-tuned models and other open-source models of similar or larger sizes across six domains: Location Navigation, Repository Management, Financial Analysis, 3D Design, Browser Automation, and Web Search. We note that most servers in the benchmark require careful configurations and thus were not included in our data synthesis pipeline. Nevertheless, \datanamen-tuned models show significant improvements on these challenging tasks compared to baselines, indicating that exposure to diverse tools enhances model performance on agentic tasks. Notably, our 32B model achieves the highest scores in 3D Design and strong performance in Financial Analysis, even outperforming much larger frontier open models like \texttt{Llama-3.3-70B-Instruct}, \texttt{Qwen2.5-72B-Instruct}, \texttt{GLM-4.5-Air} (106B), and \texttt{DeepSeek-V3} (671B).

Figure \ref{fig:mcp-universe-pareto} plots model performance versus model size on MCP-Universe benchmark. We observe that \datanamen-tuned models push the Pareto frontier forward, achieving higher OSR at smaller model sizes, indicating that \dataname can help models achieve superior performance-efficiency trade-offs in agentic tasks.

\subsection{Ablation Analysis}

\begin{wrapfigure}{r}{0.5\textwidth}
  \vspace{-1.5em} % Adjust vertical spacing above the table
  \centering
  \caption{This table shows ablation analysis of \dataname extensions.}
  \label{tab:ablation-results-table}
  \resizebox{\linewidth}{!}{ % Scale to the wrapfigure width, preserving aspect ratio
    \begin{tabular}{@{}lccc@{}}
      \toprule
      & \textbf{BFCLv3} & \multicolumn{2}{c}{\textbf{$\tau$-bench}} \\
      \cmidrule(lr){2-2} \cmidrule(lr){3-4}
      & & \textit{Airline @1} & \textit{Retail @1} \\
      \midrule
      Qwen2.5-14B-Instruct & 57.69\% & 17.25\% & 44.46\% \\
      \quad + Single Turn & 60.16\% & 15.50\% & 36.95\% \\
      \quad\quad + Irrelevance & 64.74\% & 16.75\% & 41.63\% \\
      \quad\quad\quad + Diversify & 64.56\% & 17.25\% & 43.70\% \\
      \quad\quad\quad\quad + Multi-Turn & \textbf{65.09\%} & \textbf{22.00\%} & \textbf{48.48\%} \\
      \bottomrule
    \end{tabular}
  }
  \vspace{-1.5em} % Adjust vertical spacing below the table
\end{wrapfigure}

To validate our extension designs, we perform ablation analysis on the Qwen2.5-14B-Instruct model, where we fine-tune on progressively extended versions of \datanamen, allowing us to isolate the contributions of each extension described in Section \ref{sec:toucan-extensions}. The experimental results are shown in Figrue \ref{tab:ablation-results-table}. We observe that all components contribute to improved scores. Detailed benchmark scores for the BFCL ablation study are provided in Appendix \ref{app:more-on-ablations}.

\section{Conclusion and Future Work}

This paper introduces \dataname, a tool-agentic dataset containing 1.5M trajectories designed to train better agentic models. We propose a comprehensive pipeline for data generation and demonstrate that models fine-tuned on \dataname achieve superior performance on benchmarks including BFCL-V3 and MCP-Universe.
\dataname represents the first step in a long-term effort to leverage tool use for building stronger LLM agents. Despite being a valuable contribution, we acknowledge our work exhibits certain limitations, which we plan to address through different initiatives.

\textbf{Expanding to More MCP Servers.} While our dataset is comprehensive, it was collected in June 2025, and new servers continue to emerge. We excluded MCP servers requiring special configurations (e.g., requires API keys or account setups), which simplifies the onboarding procedure but may overlook important servers and widely-used scenarios (e.g., Notion and GitHub). Manually onboarding more servers or developing automated onboarding agents could be valuable future work.

\textbf{Expert models to simulate tool-responses.} While real tool execution produces higher-quality results, it is often slow and costly, and therefore, not an option for everyone. To provide an alternative that also yields quality, we plan to develop an expert LLM capable of simulating tool execution. This artificial component will significantly reduce the cost of generating trajectory data involving tool use. Although the idea of tool-execution simulation is known within the community, it has most likely been implemented using off-the-shelf, closed-source LLMs.

\textbf{MCP Benchmark for web search.} As tool-use capabilities become central to both LLMs and LLM-agents, specific scenarios such as web search have gained prominence in the community as a means of synthesizing complex reasoning tasks. To advance this direction, we plan to develop an MCP benchmark focused on web search capabilities.
% \clearpage
% The Use of Large Language Models (LLMs)
% The use of LLMs is allowed as a general-purpose assist tool. However, new this year, if LLMs played a significant role in research ideation and/or writing to the extent that they could be regarded as a contributor, then authors should describe the precise role of the LLM in a separate section on LLM usage. This section can appear in the appendix, and will not be considered as part of the page limit. Not disclosing significant LLM usage can lead to desk rejection of the paper.

% Irrespective of the ways that LLMs were used in a given submission, authors should understand that they take full responsibility for the contents written under their name, including content generated by LLMs that could be construed as plagiarism or scientific misconduct (e.g., fabrication of facts).  LLMs are not eligible for authorship.

\section{Use of Large Language Models (LLMs)}
In our work, we used large language models (LLMs) to assist with improving the grammar, clarity, and overall readability of the manuscript, as well as to help generate the pipeline diagram included in the paper. All LLM-generated content was thoroughly verified by the authors as part of an iterative process to ensure accuracy, quality, and consistency with the scientific contributions of the work.

\section{Ethics Statement}

Developers planning to use \texttt{Toucan} for LLM fine-tuning should take into account certain considerations.

\textbf{Data Ownership and Licensing.} The MCP server specification files used to build \dataname were collected in June 2025 from \url{https://smithery.ai/}, a public platform hosting such specifications. These files were voluntarily published by their owners in accordance with the platform's privacy notice. Given the case a legitimate owner requests removal of their content from our dataset, we will honor that request through a take down process available via our GitHub repository.

\textbf{Sensitive Information.} The risk of exposing sensitive data in specification files is minimal, as they generally rely on placeholders rather than real information. However, human error may still lead to the inclusion of URLs, tokens, or email addresses. To mitigate this, we apply a pre-filtering stage with rule-based verifiers that detect common patterns of personally identifiable information (PII).  

\textbf{Data Evolution.} Our data were collected in June 2025, so \dataname captures real-world tool-use scenarios available at that time. For example, responses from search MCP servers reflect information current through June 2025. To facilitate future updates and customization, we provide our modular data pipeline, allowing researchers and practitioners to expand domain coverage and tailor tool representations for their applications.

\textbf{LLM Hallucinations.} Only tasks and annotations in \dataname were generated with LLMs; trajectories were produced using LLMs in combination with agent frameworks and remote MCP servers. This integration ensures reliable tool call executions and responses, reducing the likelihood of code errors from hallucinations. Nevertheless, hallucinations remain a general risk when using LLMs, and outputs from models fine-tuned with \dataname should always be verified by humans. 

\section{Reproducibility statement}

% It is important that the work published in ICLR is reproducible. Authors are strongly encouraged to include a paragraph-long Reproducibility Statement at the end of the main text (before references) to discuss the efforts that have been made to ensure reproducibility. This paragraph should not itself describe details needed for reproducing the results, but rather reference the parts of the main paper, appendix, and supplemental materials that will help with reproducibility. For example, for novel models or algorithms, a link to an anonymous downloadable source code can be submitted as supplementary materials; for theoretical results, clear explanations of any assumptions and a complete proof of the claims can be included in the appendix; for any datasets used in the experiments, a complete description of the data processing steps can be provided in the supplementary materials. Each of the above are examples of things that can be referenced in the reproducibility statement. This optional reproducibility statement is not part of the main text and therefore will not count toward the page limit. 

We provide the code for our data generation pipeline, along with detailed instructions for executing the pipeline end-to-end, as well as sample dataset files in the supplementary materials. The main paper and appendix further document key implementation details, including prompt templates, hyperparameter configurations used during fine-tuning, and extensions of our data analysis and fine-tuning experiments. After publication, we plan to release the full codebase in a public GitHub repository and make our datasets publicly available on the HuggingFace platform.


\bibliography{iclr2026_conference}

\begin{thebibliography}{36}
\providecommand{\natexlab}[1]{#1}
\providecommand{\url}[1]{\texttt{#1}}
\expandafter\ifx\csname urlstyle\endcsname\relax
  \providecommand{\doi}[1]{doi: #1}\else
  \providecommand{\doi}{doi: \begingroup \urlstyle{rm}\Url}\fi

\bibitem[Anthropic(2025)]{mcp-antrophic-blog}
Anthropic.
\newblock Introducing the model context protocol.
\newblock \url{https://www.anthropic.com/news/model-context-protocol}, 2025.
\newblock Accessed: 2025-08-18.

\bibitem[Barres et~al.(2025)Barres, Dong, Ray, Si, and Narasimhan]{barres2025tau2benchevaluatingconversationalagents}
Victor Barres, Honghua Dong, Soham Ray, Xujie Si, and Karthik Narasimhan.
\newblock $\tau^2$-bench: Evaluating conversational agents in a dual-control environment, 2025.
\newblock URL \url{https://arxiv.org/abs/2506.07982}.

\bibitem[Basu et~al.(2024)Basu, Abdelaziz, Chaudhury, Dan, Crouse, Munawar, Kumaravel, Muthusamy, Kapanipathi, and Lastras]{basu2024apiblendcomprehensivecorporatraining}
Kinjal Basu, Ibrahim Abdelaziz, Subhajit Chaudhury, Soham Dan, Maxwell Crouse, Asim Munawar, Sadhana Kumaravel, Vinod Muthusamy, Pavan Kapanipathi, and Luis~A. Lastras.
\newblock Api-blend: A comprehensive corpora for training and benchmarking api llms, 2024.
\newblock URL \url{https://arxiv.org/abs/2402.15491}.

\bibitem[Bousetouane(2025)]{bousetouane2025agenticsystemsguidetransforming}
Fouad Bousetouane.
\newblock Agentic systems: A guide to transforming industries with vertical ai agents, 2025.
\newblock URL \url{https://arxiv.org/abs/2501.00881}.

\bibitem[Chen et~al.(2025)Chen, Hao, Liu, Huang, Zeng, Yu, Li, Wang, Gan, Huang, Liu, Wang, Lian, Yin, Wang, and Liu]{chen_acebench_2025}
Chen Chen, Xinlong Hao, Weiwen Liu, Xu~Huang, Xingshan Zeng, Shuai Yu, Dexun Li, Shuai Wang, Weinan Gan, Yuefeng Huang, Wulong Liu, Xinzhi Wang, Defu Lian, Baoqun Yin, Yasheng Wang, and Wu~Liu.
\newblock {ACEBench}: {Who} {Wins} the {Match} {Point} in {Tool} {Usage}?, July 2025.
\newblock URL \url{http://arxiv.org/abs/2501.12851}.
\newblock arXiv:2501.12851 [cs].

\bibitem[DeepSeek-AI et~al.(2025)DeepSeek-AI, Liu, Feng, Xue, Wang, Wu, Lu, Zhao, Deng, Zhang, Ruan, Dai, Guo, Yang, Chen, Ji, Li, Lin, Dai, Luo, Hao, Chen, Li, Zhang, Bao, Xu, Wang, Zhang, Ding, Xin, Gao, Li, Qu, Cai, Liang, Guo, Ni, Li, Wang, Chen, Chen, Yuan, Qiu, Li, Song, Dong, Hu, Gao, Guan, Huang, Yu, Wang, Zhang, Xu, Xia, Zhao, Wang, Zhang, Li, Wang, Zhang, Zhang, Tang, Li, Tian, Huang, Wang, Zhang, Wang, Zhu, Chen, Du, Chen, Jin, Ge, Zhang, Pan, Wang, Xu, Zhang, Chen, Li, Lu, Zhou, Chen, Wu, Ye, Ye, Ma, Wang, Zhou, Yu, Zhou, Pan, Wang, Yun, Pei, Sun, Xiao, Zeng, Zhao, An, Liu, Liang, Gao, Yu, Zhang, Li, Jin, Wang, Bi, Liu, Wang, Shen, Chen, Zhang, Chen, Nie, Sun, Wang, Cheng, Liu, Xie, Liu, Yu, Song, Shan, Zhou, Yang, Li, Su, Lin, Li, Wang, Wei, Zhu, Zhang, Xu, Xu, Huang, Li, Zhao, Sun, Li, Wang, Yu, Zheng, Zhang, Shi, Xiong, He, Tang, Piao, Wang, Tan, Ma, Liu, Guo, Wu, Ou, Zhu, Wang, Gong, Zou, He, Zha, Xiong, Ma, Yan, Luo, You, Liu, Zhou, Wu, Ren, Ren, Sha, Fu, Xu, Huang, Zhang, Xie, Zhang, Hao,
  Gou, Ma, Yan, Shao, Xu, Wu, Zhang, Li, Gu, Zhu, Liu, Li, Xie, Song, Gao, and Pan]{deepseekai2025deepseekv3technicalreport}
DeepSeek-AI, Aixin Liu, Bei Feng, Bing Xue, Bingxuan Wang, Bochao Wu, Chengda Lu, Chenggang Zhao, Chengqi Deng, Chenyu Zhang, Chong Ruan, Damai Dai, Daya Guo, Dejian Yang, Deli Chen, Dongjie Ji, Erhang Li, Fangyun Lin, Fucong Dai, Fuli Luo, Guangbo Hao, Guanting Chen, Guowei Li, H.~Zhang, Han Bao, Hanwei Xu, Haocheng Wang, Haowei Zhang, Honghui Ding, Huajian Xin, Huazuo Gao, Hui Li, Hui Qu, J.~L. Cai, Jian Liang, Jianzhong Guo, Jiaqi Ni, Jiashi Li, Jiawei Wang, Jin Chen, Jingchang Chen, Jingyang Yuan, Junjie Qiu, Junlong Li, Junxiao Song, Kai Dong, Kai Hu, Kaige Gao, Kang Guan, Kexin Huang, Kuai Yu, Lean Wang, Lecong Zhang, Lei Xu, Leyi Xia, Liang Zhao, Litong Wang, Liyue Zhang, Meng Li, Miaojun Wang, Mingchuan Zhang, Minghua Zhang, Minghui Tang, Mingming Li, Ning Tian, Panpan Huang, Peiyi Wang, Peng Zhang, Qiancheng Wang, Qihao Zhu, Qinyu Chen, Qiushi Du, R.~J. Chen, R.~L. Jin, Ruiqi Ge, Ruisong Zhang, Ruizhe Pan, Runji Wang, Runxin Xu, Ruoyu Zhang, Ruyi Chen, S.~S. Li, Shanghao Lu, Shangyan Zhou, Shanhuang
  Chen, Shaoqing Wu, Shengfeng Ye, Shengfeng Ye, Shirong Ma, Shiyu Wang, Shuang Zhou, Shuiping Yu, Shunfeng Zhou, Shuting Pan, T.~Wang, Tao Yun, Tian Pei, Tianyu Sun, W.~L. Xiao, Wangding Zeng, Wanjia Zhao, Wei An, Wen Liu, Wenfeng Liang, Wenjun Gao, Wenqin Yu, Wentao Zhang, X.~Q. Li, Xiangyue Jin, Xianzu Wang, Xiao Bi, Xiaodong Liu, Xiaohan Wang, Xiaojin Shen, Xiaokang Chen, Xiaokang Zhang, Xiaosha Chen, Xiaotao Nie, Xiaowen Sun, Xiaoxiang Wang, Xin Cheng, Xin Liu, Xin Xie, Xingchao Liu, Xingkai Yu, Xinnan Song, Xinxia Shan, Xinyi Zhou, Xinyu Yang, Xinyuan Li, Xuecheng Su, Xuheng Lin, Y.~K. Li, Y.~Q. Wang, Y.~X. Wei, Y.~X. Zhu, Yang Zhang, Yanhong Xu, Yanhong Xu, Yanping Huang, Yao Li, Yao Zhao, Yaofeng Sun, Yaohui Li, Yaohui Wang, Yi~Yu, Yi~Zheng, Yichao Zhang, Yifan Shi, Yiliang Xiong, Ying He, Ying Tang, Yishi Piao, Yisong Wang, Yixuan Tan, Yiyang Ma, Yiyuan Liu, Yongqiang Guo, Yu~Wu, Yuan Ou, Yuchen Zhu, Yuduan Wang, Yue Gong, Yuheng Zou, Yujia He, Yukun Zha, Yunfan Xiong, Yunxian Ma, Yuting Yan, Yuxiang
  Luo, Yuxiang You, Yuxuan Liu, Yuyang Zhou, Z.~F. Wu, Z.~Z. Ren, Zehui Ren, Zhangli Sha, Zhe Fu, Zhean Xu, Zhen Huang, Zhen Zhang, Zhenda Xie, Zhengyan Zhang, Zhewen Hao, Zhibin Gou, Zhicheng Ma, Zhigang Yan, Zhihong Shao, Zhipeng Xu, Zhiyu Wu, Zhongyu Zhang, Zhuoshu Li, Zihui Gu, Zijia Zhu, Zijun Liu, Zilin Li, Ziwei Xie, Ziyang Song, Ziyi Gao, and Zizheng Pan.
\newblock Deepseek-v3 technical report, 2025.
\newblock URL \url{https://arxiv.org/abs/2412.19437}.

\bibitem[Ferrag et~al.(2025)Ferrag, Tihanyi, and Debbah]{ferrag2025llmreasoningautonomousai}
Mohamed~Amine Ferrag, Norbert Tihanyi, and Merouane Debbah.
\newblock From llm reasoning to autonomous ai agents: A comprehensive review, 2025.
\newblock URL \url{https://arxiv.org/abs/2504.19678}.

\bibitem[Gao et~al.(2025)Gao, Xie, Zhai, Ma, and Shen]{gao2025mcpradarmultidimensionalbenchmarkevaluating}
Xuanqi Gao, Siyi Xie, Juan Zhai, Shqing Ma, and Chao Shen.
\newblock Mcp-radar: A multi-dimensional benchmark for evaluating tool use capabilities in large language models, 2025.
\newblock URL \url{https://arxiv.org/abs/2505.16700}.

\bibitem[Guo et~al.(2025{\natexlab{a}})Guo, Soria, Sun, Shen, and Panda]{guo2025apipackmassivemultiprogramming}
Zhen Guo, Adriana~Meza Soria, Wei Sun, Yikang Shen, and Rameswar Panda.
\newblock Api pack: A massive multi-programming language dataset for api call generation, 2025{\natexlab{a}}.
\newblock URL \url{https://arxiv.org/abs/2402.09615}.

\bibitem[Guo et~al.(2025{\natexlab{b}})Guo, Xu, Zhu, Hong, Wang, and Mao]{guo2025mcpagentbenchevaluatingrealworldlanguage}
Zikang Guo, Benfeng Xu, Chiwei Zhu, Wentao Hong, Xiaorui Wang, and Zhendong Mao.
\newblock Mcp-agentbench: Evaluating real-world language agent performance with mcp-mediated tools, 2025{\natexlab{b}}.
\newblock URL \url{https://arxiv.org/abs/2509.09734}.

\bibitem[interstellarninja()]{Hermes-Function-Calling-Dataset-V1}
Teknium interstellarninja.
\newblock Hermes function calling dataset v1.
\newblock URL \url{https://huggingface.co/NousResearch/hermes-function-calling-v1}.

\bibitem[Li et~al.(2024)Li, Wang, Zeng, et~al.]{li2024llm}
X.~Li, S.~Wang, S.~Zeng, et~al.
\newblock A survey on llm-based multi-agent systems: workflow, infrastructure, and challenges.
\newblock \emph{Vicinagearth}, 1:\penalty0 9, 2024.
\newblock \doi{10.1007/s44336-024-00009-2}.
\newblock URL \url{https://doi.org/10.1007/s44336-024-00009-2}.

\bibitem[Liu et~al.(2025{\natexlab{a}})Liu, Huang, Zeng, Hao, Yu, Li, Wang, Gan, Liu, Yu, Wang, Wang, Ning, Hou, Wang, Wu, Wang, Liu, Wang, Tang, Tu, Shang, Jiang, Tang, Lian, Liu, and Chen]{liu2025toolacewinningpointsllm}
Weiwen Liu, Xu~Huang, Xingshan Zeng, Xinlong Hao, Shuai Yu, Dexun Li, Shuai Wang, Weinan Gan, Zhengying Liu, Yuanqing Yu, Zezhong Wang, Yuxian Wang, Wu~Ning, Yutai Hou, Bin Wang, Chuhan Wu, Xinzhi Wang, Yong Liu, Yasheng Wang, Duyu Tang, Dandan Tu, Lifeng Shang, Xin Jiang, Ruiming Tang, Defu Lian, Qun Liu, and Enhong Chen.
\newblock Toolace: Winning the points of llm function calling, 2025{\natexlab{a}}.
\newblock URL \url{https://arxiv.org/abs/2409.00920}.

\bibitem[Liu et~al.(2025{\natexlab{b}})Liu, Qiu, Wang, Zhang, Liu, Ram, Chen, Yao, Heinecke, Savarese, Wang, and Xiong]{liu2025mcpevalautomaticmcpbaseddeep}
Zhiwei Liu, Jielin Qiu, Shiyu Wang, Jianguo Zhang, Zuxin Liu, Roshan Ram, Haolin Chen, Weiran Yao, Shelby Heinecke, Silvio Savarese, Huan Wang, and Caiming Xiong.
\newblock Mcpeval: Automatic mcp-based deep evaluation for ai agent models, 2025{\natexlab{b}}.
\newblock URL \url{https://arxiv.org/abs/2507.12806}.

\bibitem[Liu et~al.(2024)Liu, Hoang, Zhang, Zhu, Lan, Kokane, Tan, Yao, Liu, Feng, Murthy, Yang, Savarese, Niebles, Wang, Heinecke, and Xiong]{liu2024apigenautomatedpipelinegenerating}
Zuxin Liu, Thai Hoang, Jianguo Zhang, Ming Zhu, Tian Lan, Shirley Kokane, Juntao Tan, Weiran Yao, Zhiwei Liu, Yihao Feng, Rithesh Murthy, Liangwei Yang, Silvio Savarese, Juan~Carlos Niebles, Huan Wang, Shelby Heinecke, and Caiming Xiong.
\newblock Apigen: Automated pipeline for generating verifiable and diverse function-calling datasets, 2024.
\newblock URL \url{https://arxiv.org/abs/2406.18518}.

\bibitem[Luo et~al.(2025)Luo, Shen, Yang, Zhao, Jwalapuram, Saha, Sahoo, Savarese, Xiong, and Li]{mcpuniverse}
Ziyang Luo, Zhiqi Shen, Wenzhuo Yang, Zirui Zhao, Prathyusha Jwalapuram, Amrita Saha, Doyen Sahoo, Silvio Savarese, Caiming Xiong, and Junnan Li.
\newblock Mcp-universe: Benchmarking large language models with real-world model context protocol servers, 2025.
\newblock URL \url{https://arxiv.org/abs/2508.14704}.

\bibitem[McInnes \& Healy(2018)McInnes and Healy]{McInnes2018UMAPUM}
Leland McInnes and John Healy.
\newblock Umap: Uniform manifold approximation and projection for dimension reduction.
\newblock \emph{ArXiv}, abs/1802.03426, 2018.
\newblock URL \url{https://api.semanticscholar.org/CorpusID:3641284}.

\bibitem[Nathawani et~al.(2025)Nathawani, Gitman, Majumdar, Bakhturina, Sunil~Mahabaleshwarkar, , Zhang, and Polak~Scowcroft]{NemotronPostTrainingDatasetV1}
Dhruv Nathawani, Igor Gitman, Somshubra Majumdar, Evelina Bakhturina, Ameya Sunil~Mahabaleshwarkar, , Jian Zhang, and Jane Polak~Scowcroft.
\newblock {Nemotron-Post-Training-Dataset-v1}, 2025.
\newblock URL \url{https://huggingface.co/datasets/nvidia/Nemotron-Post-Training-Dataset-v1}.

\bibitem[Patil et~al.(2023)Patil, Zhang, Wang, and Gonzalez]{patil2023gorillalargelanguagemodel}
Shishir~G. Patil, Tianjun Zhang, Xin Wang, and Joseph~E. Gonzalez.
\newblock Gorilla: Large language model connected with massive apis, 2023.
\newblock URL \url{https://arxiv.org/abs/2305.15334}.

\bibitem[Patil et~al.(2025)Patil, Mao, Cheng-Jie~Ji, Yan, Suresh, Stoica, and E.~Gonzalez]{patil2025bfcl}
Shishir~G. Patil, Huanzhi Mao, Charlie Cheng-Jie~Ji, Fanjia Yan, Vishnu Suresh, Ion Stoica, and Joseph E.~Gonzalez.
\newblock The berkeley function calling leaderboard (bfcl): From tool use to agentic evaluation of large language models.
\newblock In \emph{Forty-second International Conference on Machine Learning}, 2025.

\bibitem[Prabhakar et~al.(2025)Prabhakar, Liu, Zhu, Zhang, Awalgaonkar, Wang, Liu, Chen, Hoang, Niebles, Heinecke, Yao, Wang, Savarese, and Xiong]{prabhakar2025apigenmtagenticpipelinemultiturn}
Akshara Prabhakar, Zuxin Liu, Ming Zhu, Jianguo Zhang, Tulika Awalgaonkar, Shiyu Wang, Zhiwei Liu, Haolin Chen, Thai Hoang, Juan~Carlos Niebles, Shelby Heinecke, Weiran Yao, Huan Wang, Silvio Savarese, and Caiming Xiong.
\newblock Apigen-mt: Agentic pipeline for multi-turn data generation via simulated agent-human interplay, 2025.
\newblock URL \url{https://arxiv.org/abs/2504.03601}.

\bibitem[Qin et~al.(2023)Qin, Liang, Ye, Zhu, Yan, Lu, Lin, Cong, Tang, Qian, Zhao, Hong, Tian, Xie, Zhou, Gerstein, Li, Liu, and Sun]{qin2023toolllmfacilitatinglargelanguage}
Yujia Qin, Shihao Liang, Yining Ye, Kunlun Zhu, Lan Yan, Yaxi Lu, Yankai Lin, Xin Cong, Xiangru Tang, Bill Qian, Sihan Zhao, Lauren Hong, Runchu Tian, Ruobing Xie, Jie Zhou, Mark Gerstein, Dahai Li, Zhiyuan Liu, and Maosong Sun.
\newblock Toolllm: Facilitating large language models to master 16000+ real-world apis, 2023.
\newblock URL \url{https://arxiv.org/abs/2307.16789}.

\bibitem[Ren et~al.(2025)Ren, Hohman, and Moritz]{ren2025scalable}
Donghao Ren, Fred Hohman, and Dominik Moritz.
\newblock A scalable approach to clustering embedding projections, 2025.
\newblock URL \url{https://arxiv.org/abs/2504.07285}.

\bibitem[Tang et~al.(2023)Tang, Deng, Lin, Han, Liang, Cao, and Sun]{tang_toolalpaca_2023}
Qiaoyu Tang, Ziliang Deng, Hongyu Lin, Xianpei Han, Qiao Liang, Boxi Cao, and Le~Sun.
\newblock {ToolAlpaca}: {Generalized} {Tool} {Learning} for {Language} {Models} with 3000 {Simulated} {Cases}, September 2023.
\newblock URL \url{http://arxiv.org/abs/2306.05301}.
\newblock arXiv:2306.05301 [cs].

\bibitem[Team et~al.(2025{\natexlab{a}})Team, Zeng, Lv, Zheng, Hou, Chen, Xie, Wang, Yin, Zeng, Zhang, Wang, Zhong, Liu, Lu, Cao, Zhang, Huang, Wei, Cheng, An, Niu, Wen, Bai, Du, Wang, Zhu, Zhang, Wen, Wu, Xu, Huang, Zhao, Cai, Yu, Li, Ge, Huang, Zhang, Xu, Zhu, Li, Yin, Lin, Yang, Jiang, Ai, Zhu, Wang, Pan, Wang, Sun, Li, Li, Hu, Zhang, Peng, Tai, Zhang, Wang, Yang, Liu, Zhao, Liu, Yan, Liu, Chen, Li, Zhao, Ren, Jiao, Zhao, Yan, Wang, Gui, Zhao, Liu, Li, Li, Lu, Wang, Yuan, Li, Du, Du, Liu, Zhi, Gao, Wang, Yang, Xu, Fan, Wu, Ding, Wang, Zhang, Li, Xu, Zhao, Zhai, Du, Dong, Lei, Tu, Yang, Lu, Li, Li, Shuang-Li, Yang, Yi, Yu, Tian, Wang, Yu, Tam, Liang, Liu, Wang, Jia, Gu, Ling, Wang, Fan, Pan, Zhang, Zhang, Fu, Zhang, Xu, Wu, Lu, Wang, Zhou, Pan, Zhang, Wang, Li, Su, Geng, Zhu, Yang, Li, Wu, Li, Liu, Wang, Li, Zhang, Liu, Yang, Zhou, Qiao, Feng, Liu, Zhang, Wang, Yao, Wang, Liu, Chai, Li, Zhao, Chen, Zhai, Xu, Huang, Wang, Li, Dong, and Tang]{5team2025glm45agenticreasoningcoding}
5~Team, Aohan Zeng, Xin Lv, Qinkai Zheng, Zhenyu Hou, Bin Chen, Chengxing Xie, Cunxiang Wang, Da~Yin, Hao Zeng, Jiajie Zhang, Kedong Wang, Lucen Zhong, Mingdao Liu, Rui Lu, Shulin Cao, Xiaohan Zhang, Xuancheng Huang, Yao Wei, Yean Cheng, Yifan An, Yilin Niu, Yuanhao Wen, Yushi Bai, Zhengxiao Du, Zihan Wang, Zilin Zhu, Bohan Zhang, Bosi Wen, Bowen Wu, Bowen Xu, Can Huang, Casey Zhao, Changpeng Cai, Chao Yu, Chen Li, Chendi Ge, Chenghua Huang, Chenhui Zhang, Chenxi Xu, Chenzheng Zhu, Chuang Li, Congfeng Yin, Daoyan Lin, Dayong Yang, Dazhi Jiang, Ding Ai, Erle Zhu, Fei Wang, Gengzheng Pan, Guo Wang, Hailong Sun, Haitao Li, Haiyang Li, Haiyi Hu, Hanyu Zhang, Hao Peng, Hao Tai, Haoke Zhang, Haoran Wang, Haoyu Yang, He~Liu, He~Zhao, Hongwei Liu, Hongxi Yan, Huan Liu, Huilong Chen, Ji~Li, Jiajing Zhao, Jiamin Ren, Jian Jiao, Jiani Zhao, Jianyang Yan, Jiaqi Wang, Jiayi Gui, Jiayue Zhao, Jie Liu, Jijie Li, Jing Li, Jing Lu, Jingsen Wang, Jingwei Yuan, Jingxuan Li, Jingzhao Du, Jinhua Du, Jinxin Liu, Junkai Zhi, Junli
  Gao, Ke~Wang, Lekang Yang, Liang Xu, Lin Fan, Lindong Wu, Lintao Ding, Lu~Wang, Man Zhang, Minghao Li, Minghuan Xu, Mingming Zhao, Mingshu Zhai, Pengfan Du, Qian Dong, Shangde Lei, Shangqing Tu, Shangtong Yang, Shaoyou Lu, Shijie Li, Shuang Li, Shuang-Li, Shuxun Yang, Sibo Yi, Tianshu Yu, Wei Tian, Weihan Wang, Wenbo Yu, Weng~Lam Tam, Wenjie Liang, Wentao Liu, Xiao Wang, Xiaohan Jia, Xiaotao Gu, Xiaoying Ling, Xin Wang, Xing Fan, Xingru Pan, Xinyuan Zhang, Xinze Zhang, Xiuqing Fu, Xunkai Zhang, Yabo Xu, Yandong Wu, Yida Lu, Yidong Wang, Yilin Zhou, Yiming Pan, Ying Zhang, Yingli Wang, Yingru Li, Yinpei Su, Yipeng Geng, Yitong Zhu, Yongkun Yang, Yuhang Li, Yuhao Wu, Yujiang Li, Yunan Liu, Yunqing Wang, Yuntao Li, Yuxuan Zhang, Zezhen Liu, Zhen Yang, Zhengda Zhou, Zhongpei Qiao, Zhuoer Feng, Zhuorui Liu, Zichen Zhang, Zihan Wang, Zijun Yao, Zikang Wang, Ziqiang Liu, Ziwei Chai, Zixuan Li, Zuodong Zhao, Wenguang Chen, Jidong Zhai, Bin Xu, Minlie Huang, Hongning Wang, Juanzi Li, Yuxiao Dong, and Jie Tang.
\newblock Glm-4.5: Agentic, reasoning, and coding (arc) foundation models, 2025{\natexlab{a}}.
\newblock URL \url{https://arxiv.org/abs/2508.06471}.

\bibitem[Team et~al.(2025{\natexlab{b}})Team, Bai, Bao, Chen, Chen, Chen, Chen, Chen, Chen, Chen, Chen, Cui, Ding, Dong, Du, Du, Du, Du, Fan, Feng, Fu, Gao, Gao, Gao, Gao, Gu, Guan, Guo, Guo, Hu, Hao, He, He, He, Hong, Hu, Hu, Huang, Huang, Huang, Jiang, Jiang, Jin, Kang, Lai, Li, Li, Li, Li, Li, Li, Li, Li, Li, Lin, Lin, Lin, Liu, Liu, Liu, Liu, Liu, Liu, Liu, Liu, Liu, Liu, Liu, Liu, Liu, Liu, Liu, Lu, Lu, Ma, Ma, Ma, Mao, Mei, Men, Miao, Pan, Peng, Qin, Qu, Shang, Shi, Shi, Song, Su, Su, Sun, Sung, Tang, Tao, Teng, Wang, Wang, Wang, Wang, Wang, Wang, Wang, Wang, Wang, Wang, Wang, Wang, Wang, Wang, Wang, Wang, Wang, Wei, Wei, Wu, Wu, Wu, Xiao, Xie, Xiong, Xu, Xu, Xu, Xu, Xu, Xu, Xu, Xu, Xu, Xu, Yan, Yan, Yang, Yang, Yang, Yang, Yang, Yao, Yao, Ye, Ye, Yin, Yu, Yuan, Yuan, Yuan, Zhan, Zhang, Zhang, Zhang, Zhang, Zhang, Zhang, Zhang, Zhang, Zhang, Zhang, Zhang, Zhao, Zhao, Zheng, Zheng, Zhou, Zhou, Zhou, Zhu, Zhuang, and Zu]{kimiteam2025kimik2openagentic}
Kimi Team, Yifan Bai, Yiping Bao, Guanduo Chen, Jiahao Chen, Ningxin Chen, Ruijue Chen, Yanru Chen, Yuankun Chen, Yutian Chen, Zhuofu Chen, Jialei Cui, Hao Ding, Mengnan Dong, Angang Du, Chenzhuang Du, Dikang Du, Yulun Du, Yu~Fan, Yichen Feng, Kelin Fu, Bofei Gao, Hongcheng Gao, Peizhong Gao, Tong Gao, Xinran Gu, Longyu Guan, Haiqing Guo, Jianhang Guo, Hao Hu, Xiaoru Hao, Tianhong He, Weiran He, Wenyang He, Chao Hong, Yangyang Hu, Zhenxing Hu, Weixiao Huang, Zhiqi Huang, Zihao Huang, Tao Jiang, Zhejun Jiang, Xinyi Jin, Yongsheng Kang, Guokun Lai, Cheng Li, Fang Li, Haoyang Li, Ming Li, Wentao Li, Yanhao Li, Yiwei Li, Zhaowei Li, Zheming Li, Hongzhan Lin, Xiaohan Lin, Zongyu Lin, Chengyin Liu, Chenyu Liu, Hongzhang Liu, Jingyuan Liu, Junqi Liu, Liang Liu, Shaowei Liu, T.~Y. Liu, Tianwei Liu, Weizhou Liu, Yangyang Liu, Yibo Liu, Yiping Liu, Yue Liu, Zhengying Liu, Enzhe Lu, Lijun Lu, Shengling Ma, Xinyu Ma, Yingwei Ma, Shaoguang Mao, Jie Mei, Xin Men, Yibo Miao, Siyuan Pan, Yebo Peng, Ruoyu Qin, Bowen Qu, Zeyu
  Shang, Lidong Shi, Shengyuan Shi, Feifan Song, Jianlin Su, Zhengyuan Su, Xinjie Sun, Flood Sung, Heyi Tang, Jiawen Tao, Qifeng Teng, Chensi Wang, Dinglu Wang, Feng Wang, Haiming Wang, Jianzhou Wang, Jiaxing Wang, Jinhong Wang, Shengjie Wang, Shuyi Wang, Yao Wang, Yejie Wang, Yiqin Wang, Yuxin Wang, Yuzhi Wang, Zhaoji Wang, Zhengtao Wang, Zhexu Wang, Chu Wei, Qianqian Wei, Wenhao Wu, Xingzhe Wu, Yuxin Wu, Chenjun Xiao, Xiaotong Xie, Weimin Xiong, Boyu Xu, Jing Xu, Jinjing Xu, L.~H. Xu, Lin Xu, Suting Xu, Weixin Xu, Xinran Xu, Yangchuan Xu, Ziyao Xu, Junjie Yan, Yuzi Yan, Xiaofei Yang, Ying Yang, Zhen Yang, Zhilin Yang, Zonghan Yang, Haotian Yao, Xingcheng Yao, Wenjie Ye, Zhuorui Ye, Bohong Yin, Longhui Yu, Enming Yuan, Hongbang Yuan, Mengjie Yuan, Haobing Zhan, Dehao Zhang, Hao Zhang, Wanlu Zhang, Xiaobin Zhang, Yangkun Zhang, Yizhi Zhang, Yongting Zhang, Yu~Zhang, Yutao Zhang, Yutong Zhang, Zheng Zhang, Haotian Zhao, Yikai Zhao, Huabin Zheng, Shaojie Zheng, Jianren Zhou, Xinyu Zhou, Zaida Zhou, Zhen Zhu,
  Weiyu Zhuang, and Xinxing Zu.
\newblock Kimi k2: Open agentic intelligence, 2025{\natexlab{b}}.
\newblock URL \url{https://arxiv.org/abs/2507.20534}.

\bibitem[Team(2024)]{qwen2.5}
Qwen Team.
\newblock Qwen2.5: A party of foundation models, September 2024.
\newblock URL \url{https://qwenlm.github.io/blog/qwen2.5/}.

\bibitem[Team(2025{\natexlab{a}})]{mcpmark_2025}
The~MCPMark Team.
\newblock Mcpmark: Stress-testing comprehensive mcp use.
\newblock \url{https://github.com/eval-sys/mcpmark}, 2025{\natexlab{a}}.

\bibitem[Team(2025{\natexlab{b}})]{scale2025mcp}
The Scale~Research Team.
\newblock Actions, not words: Mcp-atlas raises the bar for agentic evaluation.
\newblock \url{https://scale.com/blog/mcp-atlas}, September 2025{\natexlab{b}}.
\newblock Accessed: YYYY-MM-DD.

\bibitem[Wang et~al.(2025)Wang, Chang, Patel, Biju, Wu, Liu, Ding, Rezazadeh, Shah, Bao, and Siow]{wang2025mcpbenchbenchmarkingtoolusingllm}
Zhenting Wang, Qi~Chang, Hemani Patel, Shashank Biju, Cheng-En Wu, Quan Liu, Aolin Ding, Alireza Rezazadeh, Ankit Shah, Yujia Bao, and Eugene Siow.
\newblock Mcp-bench: Benchmarking tool-using llm agents with complex real-world tasks via mcp servers, 2025.
\newblock URL \url{https://arxiv.org/abs/2508.20453}.

\bibitem[Xu et~al.(2025{\natexlab{a}})Xu, Huang, Gao, and Shang]{xu2025llm}
Weikai Xu, Chengrui Huang, Shen Gao, and Shuo Shang.
\newblock Llm-based agents for tool learning: A survey.
\newblock \emph{Data Science and Engineering}, 2025{\natexlab{a}}.
\newblock \doi{10.1007/s41019-025-00296-9}.
\newblock URL \url{https://link.springer.com/article/10.1007/s41019-025-00296-9}.

\bibitem[Xu et~al.(2025{\natexlab{b}})Xu, Liu, Yin, Zhou, and Poovendran]{Xu2025KodCodeAD}
Zhangchen Xu, Yang Liu, Yueqin Yin, Mingyuan Zhou, and Radha Poovendran.
\newblock Kodcode: A diverse, challenging, and verifiable synthetic dataset for coding.
\newblock \emph{ArXiv}, abs/2503.02951, 2025{\natexlab{b}}.
\newblock URL \url{https://api.semanticscholar.org/CorpusID:276782338}.

\bibitem[Yan et~al.(2025)Yan, Wang, Du, Yang, Shan, Qiu, Jia, Wang, Yuan, Han, Qin, Chen, Peng, Wang, and Xu]{yan2025mcpworldunifiedbenchmarkingtestbed}
Yunhe Yan, Shihe Wang, Jiajun Du, Yexuan Yang, Yuxuan Shan, Qichen Qiu, Xianqing Jia, Xinge Wang, Xin Yuan, Xu~Han, Mao Qin, Yinxiao Chen, Chen Peng, Shangguang Wang, and Mengwei Xu.
\newblock Mcpworld: A unified benchmarking testbed for api, gui, and hybrid computer use agents, 2025.
\newblock URL \url{https://arxiv.org/abs/2506.07672}.

\bibitem[Yang et~al.(2025)Yang, Li, Yang, Zhang, Hui, Zheng, Yu, Gao, Huang, Lv, Zheng, Liu, Zhou, Huang, Hu, Ge, Wei, Lin, Tang, Yang, Tu, Zhang, Yang, Yang, Zhou, Zhou, Lin, Dang, Bao, Yang, Yu, Deng, Li, Xue, Li, Zhang, Wang, Zhu, Men, Gao, Liu, Luo, Li, Tang, Yin, Ren, Wang, Zhang, Ren, Fan, Su, Zhang, Zhang, Wan, Liu, Wang, Cui, Zhang, Zhou, and Qiu]{yang2025qwen3technicalreport}
An~Yang, Anfeng Li, Baosong Yang, Beichen Zhang, Binyuan Hui, Bo~Zheng, Bowen Yu, Chang Gao, Chengen Huang, Chenxu Lv, Chujie Zheng, Dayiheng Liu, Fan Zhou, Fei Huang, Feng Hu, Hao Ge, Haoran Wei, Huan Lin, Jialong Tang, Jian Yang, Jianhong Tu, Jianwei Zhang, Jianxin Yang, Jiaxi Yang, Jing Zhou, Jingren Zhou, Junyang Lin, Kai Dang, Keqin Bao, Kexin Yang, Le~Yu, Lianghao Deng, Mei Li, Mingfeng Xue, Mingze Li, Pei Zhang, Peng Wang, Qin Zhu, Rui Men, Ruize Gao, Shixuan Liu, Shuang Luo, Tianhao Li, Tianyi Tang, Wenbiao Yin, Xingzhang Ren, Xinyu Wang, Xinyu Zhang, Xuancheng Ren, Yang Fan, Yang Su, Yichang Zhang, Yinger Zhang, Yu~Wan, Yuqiong Liu, Zekun Wang, Zeyu Cui, Zhenru Zhang, Zhipeng Zhou, and Zihan Qiu.
\newblock Qwen3 technical report, 2025.
\newblock URL \url{https://arxiv.org/abs/2505.09388}.

\bibitem[Yao et~al.(2024)Yao, Shinn, Razavi, and Narasimhan]{yao2024taubenchbenchmarktoolagentuserinteraction}
Shunyu Yao, Noah Shinn, Pedram Razavi, and Karthik Narasimhan.
\newblock $\tau$-bench: A benchmark for tool-agent-user interaction in real-world domains, 2024.
\newblock URL \url{https://arxiv.org/abs/2406.12045}.

\bibitem[Yin et~al.(2025)Yin, Shen, Xu, Han, Dong, Zhang, Hu, Liu, Ma, Wang, et~al.]{yin2025livemcp}
Ming Yin, Dinghan Shen, Silei Xu, Jianbing Han, Sixun Dong, Mian Zhang, Yebowen Hu, Shujian Liu, Simin Ma, Song Wang, et~al.
\newblock Livemcp-101: Stress testing and diagnosing mcp-enabled agents on challenging queries.
\newblock \emph{arXiv preprint arXiv:2508.15760}, 2025.

\end{thebibliography}
\bibliographystyle{iclr2026_conference}

\appendix
\newpage

\clearpage
\section{Dataset Schema and Examples}
\label{app:dataset-schema}

An instance of \dataname contains the following columns:
\begin{itemize}
    \item \textbf{uuid:} Unique sample identifier.
    \item \textbf{subset:} Annotation specifying which pipeline was used to generate the trajectory. Options:
    (1) \textit{single-turn-original:} only the core processing (Stage 1 to 5) described in Section~\ref{sec:toucan-overview} are applied, (2) \textit{irrelevant:} a server shuffle process applied on top of the \textit{single-turn-original} pipeline, (3) \textit{single-turn-diversify:} a question diversification process applied on top of the \textit{single-turn-original} pipeline, and (4) \textit{multi-turn:} a multi-turn extension of the \textit{single-turn-original} and \textit{single-turn-diversify} subsets.
    \item \textbf{messages:} The trajectory formatted with the chat template from the original LLM-agent used for generation. The system prompt includes the associated list of tools.
    \item \textbf{question:} The user task crafted to generate the trajectory.
    \item \textbf{target\_tools:} The MCP tools used as seeds for question generation.
    \item \textbf{question\_quality\_assessment:} Task evaluation by an LLM-as-judge, covering quality, difficulty, realism, and uniqueness.
    \item \textbf{response\_quality\_assessment:} Response evaluation by an LLM-as-judge, covering completeness and conciseness.
    \item \textbf{message\_num\_rounds:} Total number of messages, including turns of all types.
    \item \textbf{metadata:} Original MCP server data collected and used as seed for generation, as well as respective LLM annotations.
\end{itemize}

This is the structure of an instance in \dataname:
\begin{minted}
[fontsize=\footnotesize,
 breaklines,
 breakanywhere]
{json}
    {
      "uuid": "3ac8fdcc-b9b5-50d2-a840-947a42b558d2",
      "subset": "single-turn-original",
      "messages": "[{...long JSON string of messages...}]",
      "question": "I'm reviewing my company's policy document that needs quarterly updates for compliance purposes. I need to first understand its current structure with all sections and subsections, then update every occurrence of 'Q3 2024' to 'Q1 2025' throughout the document, and finally add explanatory footnotes to the last 3 sections that reference our new regulatory guidance document published last month. Can you help me accomplish these updates systematically?",
      "target_tools": "get_document_outline, search_and_replace, add_footnote_to_document",
      "question_quality_assessment_kimik2w4a16": {
        "tool_selection_difficulty": { "reasoning": "...", "score": 3 },
        "tool_selection_uniqueness": { "reasoning": "...", "score": 3 },
        "question_quality": { "reasoning": "...", "score": 4 },
        "scenario_realism": { "reasoning": "...", "score": 5 },
        "verifiable": { "reasoning": "...", "score": 4 },
        "stability": { "reasoning": "...", "score": 5 },
        "overall_score": 4.0
      },
      "response_quality_assessment_oss120b": {
        "completeness": { "reasoning": "...", "score": 2 },
        "conciseness": { "reasoning": "...", "score": 3 },
        "overall_score": 2.5,
        "desired_tools_used_percentage": 0.3333,
        "order_correctness": false
      },
      "messages_num_rounds": 9,
      "metadata": "{...long metadata JSON string...}"
    }
\end{minted}

Below, we present a \dataname trajectory, chat template tokens from the model generator and emojis have been removed, and some sections (e.g., tool list, tool output, assistant response) have been abbreviated to improve readability.

\label{app:trajectory-example}
\begin{minted}
[fontsize=\footnotesize,
 breaklines,
 breakanywhere]
{json}
[
  {
    "role": "system",
    "content": "{... contains the list of tools ... }"
  },
  {
    "role": "user",
    "content": "Can you explain what ENS (Ethereum Name Service) represents in the context of a transaction and why a name like 'blockscout.eth' might appear instead of a raw address?"
  },
  {
    "role": "assistant",
    "content": "I'll explain what ENS (Ethereum Name Service) represents, and show you how it resolves a name like 'blockscout.eth' into its actual Ethereum address."
  },
  {
    "role": "assistant",
    "content": "",
    "function_call": {
      "name": "blockscout-mcp-server-__unlock_blockchain_analysis__",
      "arguments": "{}"
    }
  },
  {
    "role": "function",
    "content": "{... long tool output ...}"
  },
  {
    "role": "assistant",
    "content": "Now, let me resolve 'blockscout.eth' to show the actual address it represents."
  },
  {
    "role": "assistant",
    "content": "",
    "function_call": {
      "name": "blockscout-mcp-server-get_address_by_ens_name",
      "arguments": "{\"name\": \"blockscout.eth\"}"
    }
  },
  {
    "role": "function",
    "name": "blockscout-mcp-server-get_address_by_ens_name",
    "content": "{\n  \"data\": {\n    \"resolved_address\": \"0x441e1e47a6Fa2DBfD3cd9b54291E9AB3a58D7975\"\n  },\n  \"data_description\": null,\n  \"notes\": null,\n  \"instructions\": null,\n  \"pagination\": null\n}"
  },
  {
    "role": "assistant",
    "content": "{... parsed assistant response ...}"
  },
  {
    "role": "user",
    "content": "Based on that explanation, could you look up the ENS name 'blockscout.eth' and provide the Ethereum address it resolves to?"
  },
  {
    "role": "assistant",
    "content": "I already resolved that for you in my previous response! When I called the ENS lookup tool, `\"blockscout.eth\"` resolved to the Ethereum address:\n\n`0x441e1e47a6Fa2DBfD3cd9b54291E9AB3a58D7975`"
  }
]
\end{minted}

In the previous trajectory example, we abbreviated the assistant's parsed response for readability. The full response is shown below.

\label{app:parsed-assistant-resonse}
\begin{minted}[fontsize=\footnotesize, linenos, breaklines, breakanywhere]{markdown}
## What ENS (Ethereum Name Service) Represents in Transactions

ENS stands for **Ethereum Name Service**, and it functions much like "DNS for Ethereum addresses." It allows users to register human-readable names (like "blockscout.eth") that map to complex hexadecimal blockchain addresses.

### Key Points About ENS:

1. **Human-Readable Names for Complex Addresses**
   - Instead of displaying `0x441e1e47a6Fa2DBfD3cd9b54291E9AB3a58D7975` (which is the raw address for blockscout.eth), interfaces show the friendly "blockscout.eth"
   - This reduces errors from copy/pasting long hex addresses

2. **Reverse Resolution**
   - When you see "blockscout.eth" in transaction data, wallet labels, or block explorers, it means:
     - ENS has been configured with this name
     - The system has performed a reverse lookup to find the name associated with the address

3. **Self-Sovereign Identity**
   - ENS names are owned and controlled by users through smart contracts
   - They can be transferred, updated, or associated with additional metadata (social profiles, avatars, etc.)

### Why "blockscout.eth" Appears Instead of Raw Address:

- **User Experience**: "blockscout.eth" is immediately recognizable and memorable
- **Trust Verification**: Users can verify they're interacting with the legitimate Blockscout entity rather than a random address
- **Branding**: Organizations use ENS names for consistent branding across DeFi protocols and wallet interfaces
- **Error Prevention**: Reduces costly mistakes when users verify transaction recipients

Every time you see an ENS name like "blockscout.eth" in a transaction, the underlying system has actually resolved this to the precise blockchain address `0x441e1e47a6Fa2DBfD3cd9b54291E9AB3a58D7975`-the ENS simply provides a human-friendly layer on top of the raw cryptographic identity.
\end{minted}
\newpage
\section{More on Dataset Analysis}
\label{app:more-data-analysis}

\subsection{Featured Servers in \dataname}
\label{app:featured-server-list}

Table \ref{tab:server_info} demonstrates featured servers of \datanamen. We note that these featured servers span a wide range of domains, making combinations possible for complex tool use tasks.

\begin{table}[h]
\centering
\resizebox{\textwidth}{!}{
\begin{tabular}{l l l c}
\toprule
\textbf{Server Name} & \textbf{Repository Name} & \textbf{Domain} & \textbf{Tools Count} \\
\midrule
OKX Server & esshka/okx-mcp & Cryptocurrency \& Blockchain & 2 \\
AI Research Assistant - Semantic Scholar & Access via Smithery \footnote{http://lit-review-assistant.streamlit.app/} & Web Search \& Research & 10 \\
Book Search Server & Access via Smithery \footnote{https://smithery.ai/server/@Busra-ozer/book-search-mcp} & Web Search \& Research & 1 \\
PubMed MCP Server & {JackKuo666/PubMed-MCP-Server} & Web Search \& Research & 4 \\
Flux ImageGen Server & {falahgs/flux-imagegen-mcp-server} & AI/ML Tools & 3 \\
Pokémcp & {NaveenBandarage/poke-mcp} & Data Analysis \& Processing & 4 \\
Hotel Booking Server & {jinkoso/jinko-mcp} & E-commerce & 6 \\
Cloudflare Playwright & {cloudflare/playwright-mcp} & Browser Automation & 24 \\
Time MCP Server & {yokingma/time-mcp} & Time \& Calendar & 6 \\
Exa Search & {exa-labs/exa-mcp-server} & Web Search \& Research & 8 \\
Weather Forecast Server & {iremaltunay55/deneme} & Weather & 5 \\
Advanced Calculator Server & {alan5543/calculator-mcp} & Data Analysis \& Processing & 17 \\
Dictionary Server & {ceydasimsekk/dictionarymcp} & Others & 1 \\
Airbnb Search and Listing Details Server & {AkekaratP/mcp-server-airbnb} & Web Search \& Research & 2 \\
Code Runner MCP Server & {formulahendry/mcp-server-code-runner} & Development Tools & 1 \\
Movie Recommender & {iremert/movie-recommender-mcp} & Content Creation & 1 \\
United States Weather & {smithery-ai/mcp-servers} & Weather & 6 \\
Context7 & {upstash/context7-mcp} & Development Tools & 2 \\
Think Tool Server & {PhillipRt/think-mcp-server} & Memory Management & 1 \\
OpenAPI MCP Server & {janwilmake/openapi-mcp-server} & API Integration & 2 \\
Film Information Server & {zehranurugurr/film\_mcp} & Content Creation & 1 \\
Trends Hub & {baranwang/mcp-trends-hub} & News \& Media & 21 \\
ClinicalTrials MCP Server & {JackKuo666/ClinicalTrials-MCP-Server} & Health \& Fitness & 7 \\
Drawing Tool for AI Assistants & {flrngel/mcp-painter} & Content Creation & 4 \\
LeetCode & {jinzcdev/leetcode-mcp-server} & Development Tools & 9 \\
\bottomrule
\end{tabular}
}
\caption{Featured Server Information}
\label{tab:server_info}
\end{table}

\subsection{More on MCP Server Analysis in \datanamen}
\label{app:top-mcps}

\begin{figure}[!h]
    \centering
    \includegraphics[width=0.8\linewidth]{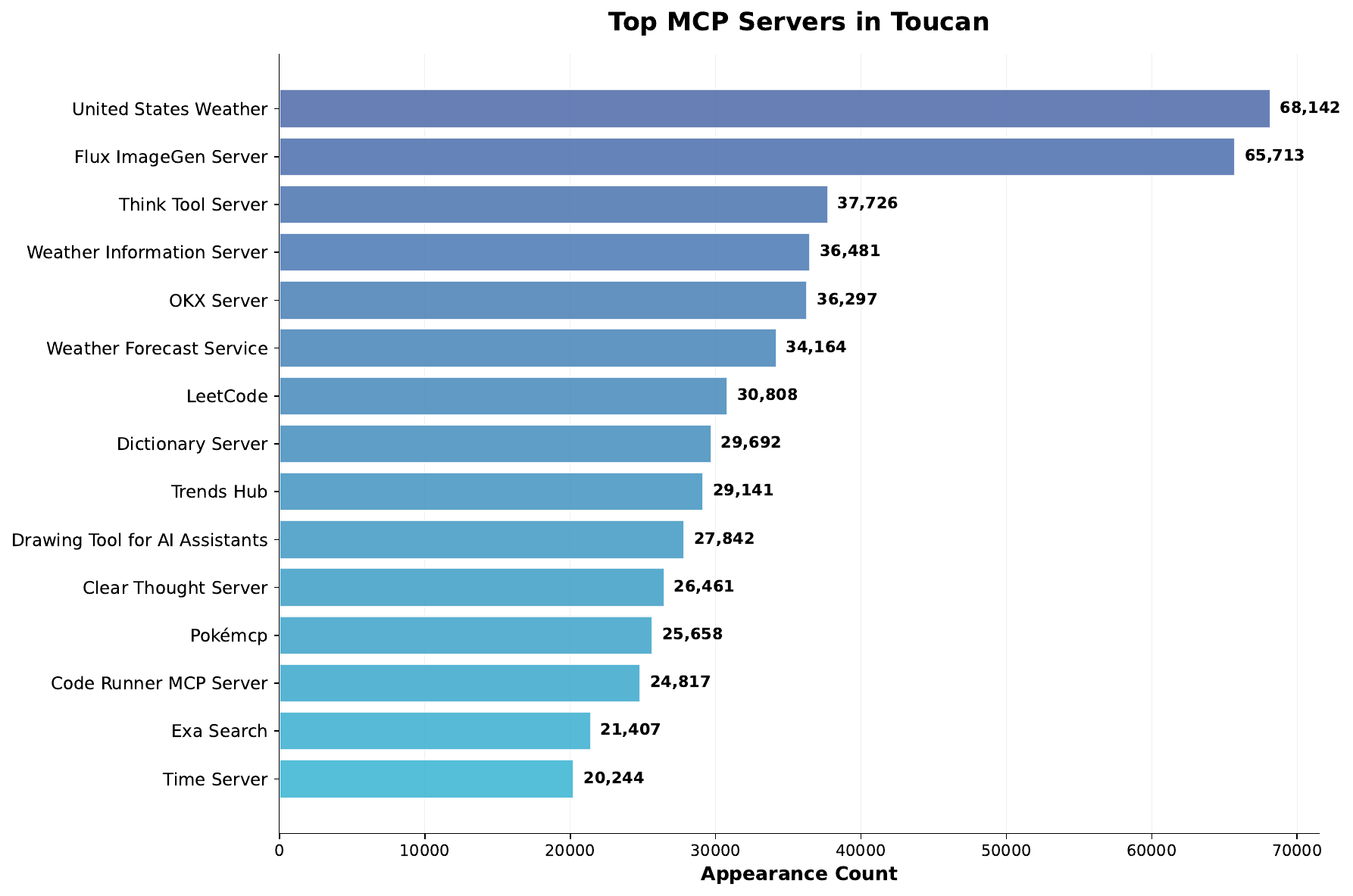}
    \caption{Distribution of the most frequently occurring MCP servers in the \dataname dataset.}
    \label{fig:top-mcp-servers}
\end{figure}

\begin{figure}[!h]
    \centering
    \includegraphics[width=0.4\linewidth]{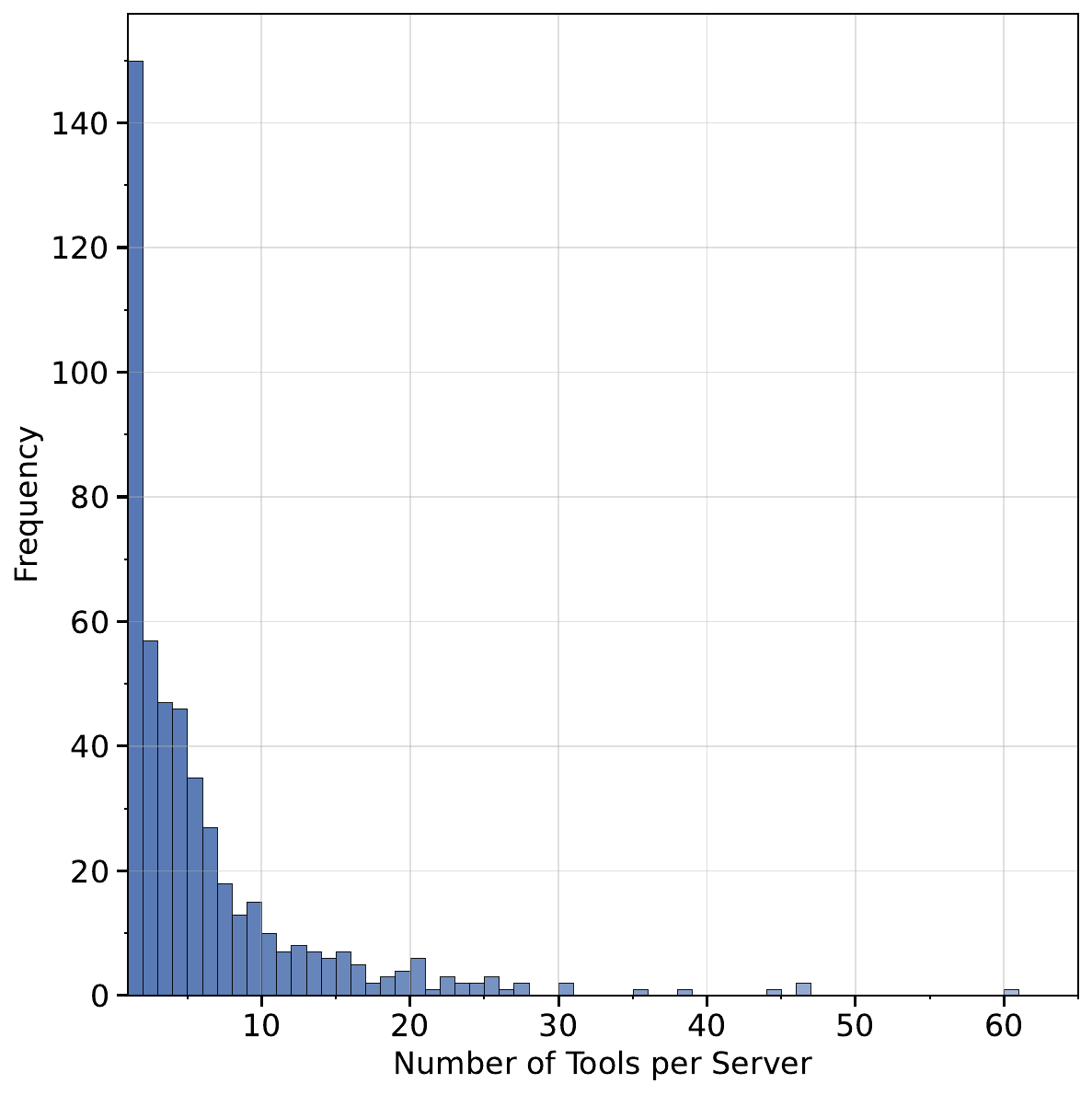}
    \caption{Tools Number distribution across MCP servers}
    \label{fig:tools-per-server}
\end{figure}

Figure~\ref{fig:top-mcp-servers} shows the distribution of the most frequently used MCP servers in our dataset, highlighting the diversity of servers and domains covered in \datanamen. Figure~\ref{fig:tools-per-server} shows the distribution of tool counts across the 495 MCP servers employed by \dataname, revealing that most servers expose only a limited number of tools, with the majority containing fewer than 10 tools.

\subsection{Emedding Visualization}
\label{app:embedding-visualization}

\begin{figure}[!h]
    \centering
    \includegraphics[width=1.0\linewidth]{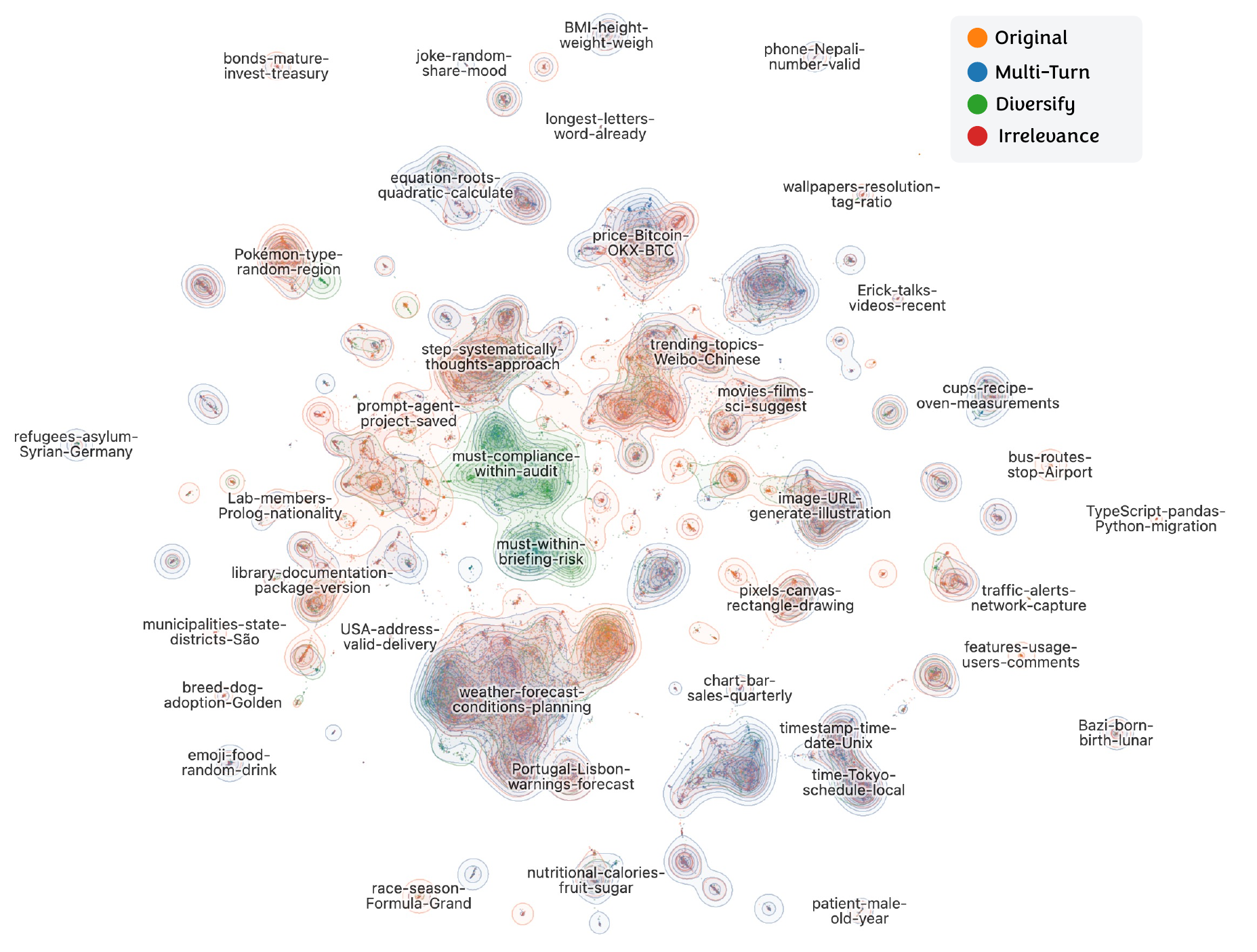}
    \caption{This figure is the visualization of 50K random-sampled \dataname instances via Embedding Atlas \citep{ren2025scalable}.}
    \label{fig:mcp-servers-category-analysis}
\end{figure}

Figure~\ref{fig:mcp-servers-category-analysis} presents embedding visualization via Embedding Atlas \citep{ren2025scalable} using the \texttt{Xenova/multilingual-e5-small} embedding model with UMAP projection \cite{McInnes2018UMAPUM}. The visualization demonstrates that \dataname covers a wide range of topics. In addition, the proposed \dataname extensions (e.g., diversification) effectively increase the overall dataset coverage.

\clearpage

\begin{wrapfigure}{r}{0.45\textwidth}
  \vspace{-6em}
  \centering
  \includegraphics[width=0.44\textwidth]{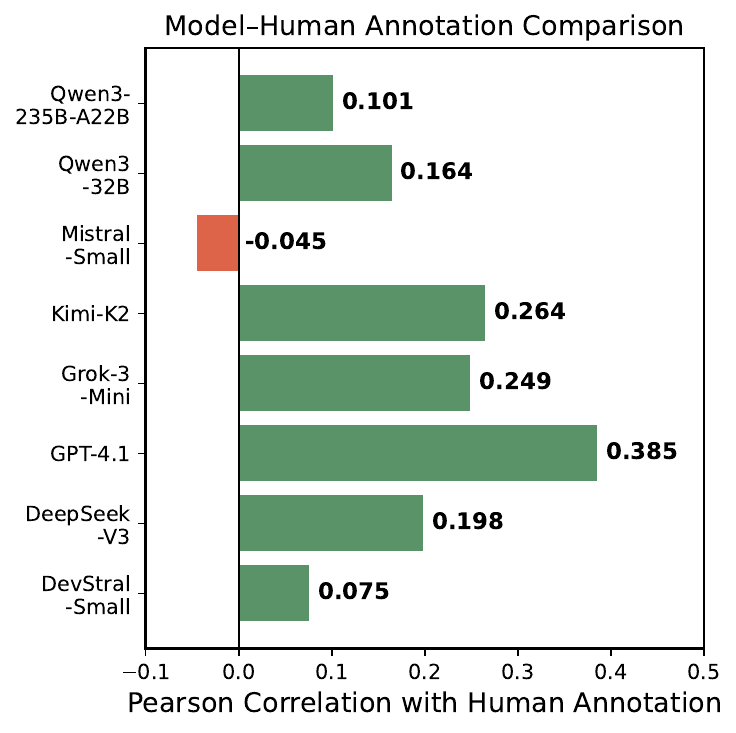}
  \caption{Pearson correlation between human annotators and LLM-as-a-Judge evaluations across different models.}
  \label{fig:annotation-correlation}
  \vspace{-3em}
\end{wrapfigure}

\section{More on Experiments}

\subsection{LLM Annotation}
\label{app:llm-annotation}

Figure~\ref{fig:annotation-correlation} shows the Pearson correlation between human annotators and LLM-as-a-Judge evaluations across different models, based on 50 randomly sampled instances. The annotation prompt is available in Appendix \ref{app:task-annotation-prompt-for-single-server}. We observe that \texttt{GPT-4.1} and \texttt{Kimi-K2} achieve the highest overall correlation with human judgments. Considering cost efficiency, we deploy \texttt{Kimi-K2} locally for our annotation pipeline.

\subsection{Fine-Tuning Hyper-Parameters}\label{app:fine-tune-parameter}

We fine-tune models with \dataname using a super computing cluster, which is outfitted with NVIDIA H100 GPUs. The fine-tuning hyper-parameters can be found in Table \ref{tab:fine-tune hyperparameters}.

\begin{table}[htbp]
\small
\centering
% \vspace{-1em}
\caption{This table shows the hyper-parameters for supervised fine-tuning.}
% \vspace{1em}
\begin{tabular}{ll}
\toprule
\textbf{Hyper-parameter} & \textbf{Value} \\ \midrule
Tool-Call Template & Hermes \\
Learning Rate & $2 \times 10^{-5}$ \\
Number of Epochs & $2$ \\
Number of Devices & $8$ or $64$ \\
Per-device Batch Size & $1$ \\
Gradient Accumulation Steps & $8$ (8 GPUs) or $1$ (64 GPUs) \\
Effective Batch Size & $64$ \\
Optimizer & \texttt{Adamw} with $\beta s=(0.9,0.999)$ and $\epsilon=10^{-8}$\\
Deepspeed & zero3 \\
Max Sequence Length  & $32768$ \\ \bottomrule
\end{tabular}
% \vspace{-0.5em}
\label{tab:fine-tune hyperparameters}
\end{table}

\subsection{More on Ablation Studies}
\label{app:more-on-ablations}

Table~\ref{app:bfclv3-results-table-ablation} details the individual scores of the BFCL V3 benchmark for our ablation analysis. We observe that all extensions are meaningful in improving model performance.

\begin{table}[h!]
  \centering
  \caption{Ablation of \dataname Extensions on BFCL V3 Benchmark.}
  \label{app:bfclv3-results-table-ablation}
  \resizebox{\textwidth}{!}{%
  \begin{tabular}{@{}lllllll@{}}
\toprule
                                                     & \textbf{Overall} & \multicolumn{2}{c}{\textbf{Single Turn}} & \textbf{Multi Turn} & \multicolumn{2}{c}{\textbf{Hallucination}} \\
                                                     &                  & \textit{Non-live (AST)}        & \textit{Live (AST)}       &                     & \textit{Relevance}           & \textit{Irrelevance}          \\
                                                 \midrule
Qwen2.5-14B-Instruct                            & 57.69\%          & 83.38\%               & 73.70\%          & 19.75\%             & 83.33\%             & 68.46\%              \\
\quad+ Single Turn                                        & 60.16\%          & 87.50\%               & 66.86\%          & 34.38\%             & 72.22\%             & 46.88\%              \\
\quad\quad+ Irrelevance                          & 64.74\%          & 88.46\%               & 77.25\%          & 30.38\%             & 72.22\%             & 77.85\%              \\
\quad\quad\quad+ Diversify              & 64.56\%          & 86.06\%               & 76.90\%          & 32.50\%             & 72.22\%             & 75.45\%              \\
\quad\quad\quad\quad+ Multi-Turn & 65.09\%          & 85.42\%               & 76.01\%          & 35.25\%             & 72.22\%             & 75.96\%              \\ \bottomrule
\end{tabular}
  }
\end{table}
% \end{document}

\clearpage
\section{Prompts}
\label{app:prompts}

\subsection{MCP Server Annotation Prompt}
\label{app:server-category-annotation-prompt}

Below is the prompt for annotating MCP server categories.

\label{app:mcp-server-annotation-prompt}
\begin{minted}[fontsize=\footnotesize, linenos, breaklines, breakanywhere]{markdown}
## Task
Generate **Server Labels** to categorize the provided MCP Server based on its description and available tools.

## Objective
Analyze the provided MCP Server's description and available tools, then assign appropriate category labels that best describe its primary functionality and use cases.

## Guidelines

### Label Selection
- Analyze the MCP Server's core functionality and purpose
- Consider the types of tools it provides and the problems it solves
- Select labels that accurately represent the server's primary use cases
- Choose from predefined categories when applicable, but also consider custom labels for unique functionality

### Predefined Categories
Choose from these established categories when appropriate:
- **Web Search & Research**: Tools for searching the web, gathering information, academic research
- **Browser Automation**: Web scraping, automated browsing, page interaction
- **Memory Management**: Data storage, retrieval, knowledge bases, note-taking
- **Operating System**: File operations, system commands, process management
- **Data Analysis & Processing**: Analytics, data transformation, statistical analysis
- **Cryptocurrency & Blockchain**: Trading, wallet management, DeFi, blockchain interaction
- **Daily Productivity**: Task management, scheduling, personal organization
- **File Management**: File operations, document handling, storage management
- **Database Operations**: Data querying, database management, SQL operations
- **API Integration**: Third-party service integration, webhook handling
- **Communication Tools**: Messaging, email, notifications, social interaction
- **Development Tools**: Code analysis, debugging, version control, CI/CD
- **Security & Authentication**: Password management, encryption, access control
- **Cloud Services**: Cloud platform integration, serverless functions
- **AI/ML Tools**: Machine learning, model interaction, AI-powered features
- **Content Creation**: Writing, editing, media generation, publishing
- **Social Media**: Social platform integration, posting, analytics
- **Financial Services**: Banking, payments, financial data, accounting
- **E-commerce**: Shopping, product management, order processing
- **Gaming**: Game-related tools, entertainment, interactive features
- **Education**: Learning tools, course management, educational content
- **Health & Fitness**: Health monitoring, fitness tracking, medical tools
- **Travel & Maps**: Location services, travel planning, navigation
- **News & Media**: News aggregation, media consumption, journalism tools
- **Weather**: Weather data, forecasting, climate information
- **Time & Calendar**: Scheduling, time management, calendar integration

### Custom Labels
- If the server doesn't fit well into predefined categories, create a custom label
- Custom labels should be descriptive and specific to the server's unique functionality
- Use clear, concise terminology that would be useful for clustering and organization

### Output Requirements
- **Primary Label**: The main category that best describes the server (from predefined list or custom)
- **Secondary Labels**: Additional relevant categories (0-2 labels)
- **Custom Label**: A free-form descriptive label if the server has unique functionality not covered by predefined categories

## MCP Server Description
{MCP_SERVER_NAME}: {MCP_SERVER_DESCRIPTION}

Available Tools:
{TOOL_LIST}

## Output
Provide your response in the following XML format:

<response>
  <analysis>
    <!-- Briefly analyze the MCP Server's core functionality and the types of problems it solves based on its description and available tools. -->
  </analysis>
  <reasoning>
    <!-- Brief explanation of why these labels were chosen and how they represent the server's functionality -->
  </reasoning>
  <primary_label>
    <!-- The main category that best describes this server's primary functionality -->
  </primary_label>
  <secondary_labels>
    <!-- Additional relevant categories (0-2 labels), separated by commas if multiple -->
  </secondary_labels>
  <custom_label>
    <!-- A free-form descriptive label if the server has unique functionality not covered by predefined categories. Leave empty if not needed. -->
  </custom_label>
</response>
\end{minted}

\subsection{Task Generation Prompt}
\label{app:task-generation-prompt-for-single-server}

Below is an example of a task generation prompt for the single-server task synthesis. The prompt generates a question targeting \textbf{one tool}.

\begin{minted}[fontsize=\footnotesize, linenos, breaklines, breakanywhere]{markdown}
## Task
Generate a **Tool Use Question** based on the provided MCP Server and its tool descriptions.

## Objective
Analyze the provided MCP Server and its available tools, then create a realistic user question that would naturally require the use of one of these tools to solve.

## Guidelines

### Question Realism
- Create questions that represent real-world scenarios where users would need to interact with the MCP Server's tools
- The question should sound natural and authentic, as if asked by someone genuinely needing to accomplish a task
- Consider common use cases, problems, or workflows that would require the functionality provided by the MCP Server's tools

### Tool Selection
- Focus on **ONE specific tool** from the MCP Server that would be most appropriate to answer the question
- Choose tools based on the core functionality they provide and how they would solve real user problems
- Consider each tool's description and purpose when crafting the question

### Question Complexity
- Create questions that are clear and specific enough to warrant tool usage
- Avoid overly simple questions that could be answered without tools
- Include relevant context or constraints that make the tool usage necessary
- Do not contain the exact tool name in the question

### Output Format
Your response should include:
1. **Tool Analysis**: Briefly analyze the MCP Server's available tools and their main functionalities.
2. **Target Tool**: The specific tool name from the MCP Server that should be used to answer this question.
3. **Question**: A clear, realistic user question that requires tool usage.

## MCP Server Description
{MCP_SERVER_NAME}: {MCP_SERVER_DESCRIPTION}

Available Tools:
{TOOL_LIST}

## Output
Provide your response in the following XML format:

<response>
  <server_analysis>
    <!-- Briefly analyze the MCP Server's available tools and their main functionalities. -->
  </server_analysis>
  <target_tool>
    <!-- The specific tool name from the MCP Server that should be used to answer this question. -->
  </target_tool>
  <question>
    <!-- A clear, realistic user question that requires tool usage. -->
  </question>
</response>
\end{minted}

Below is an example of a task generation prompt for the single-server task synthesis. The prompt generates a question targeting \textbf{multiple tools}.

\begin{minted}[fontsize=\footnotesize, linenos, breaklines, breakanywhere]{markdown}
## Task
Generate a **Tool Use Question** based on the provided MCP Server and its tool descriptions.

## Objective
Analyze the provided MCP Server and its available tools, then create a realistic user question that would naturally require the use of **{NUM_TOOLS} tools** from this MCP Server to solve completely.

## Guidelines

### Question Realism
- Create questions that represent real-world scenarios where users would need to interact with the MCP Server's tools
- The question should sound natural and authentic, as if asked by someone genuinely needing to accomplish a task
- Consider common use cases, problems, or workflows that would require the functionality provided by the MCP Server's tools

### Tool Selection
- Focus on **{NUM_TOOLS} tools** from the MCP Server that would work together to answer the question
- The question should require a sequence or combination of tool calls to solve completely
- Choose tools based on how they complement each other and create a logical workflow
- Consider each tool's description and purpose when crafting the question that requires multiple steps

### Question Complexity
- Create questions that are complex enough to warrant using {NUM_TOOLS} tools
- The question should have multiple components or require several steps to solve
- Include relevant context or constraints that make the multi-tool usage necessary
- Do not contain the exact tool names in the question
- Ensure the question cannot be reasonably answered with just a single tool

### Output Format
Your response should include:
1. **Tool Analysis**: Briefly analyze the MCP Server's available tools and their main functionalities.
2. **Target Tools**: The specific tool names from the MCP Server that should be used together to answer this question, in the order they would likely be called.
3. **Question**: A clear, realistic user question that requires multiple tool usage.

## MCP Server Description
{MCP_SERVER_NAME}: {MCP_SERVER_DESCRIPTION}

Available Tools:
{TOOL_LIST}

## Output
Ensure your question requires exactly {NUM_TOOLS} tools to solve completely. Provide your response in the following XML format:

<response>
  <server_analysis>
    <!-- Briefly analyze the MCP Server's available tools and their main functionalities. -->
  </server_analysis>
  <target_tools>
    <!-- The specific tool names from the MCP Server that should be used together to answer this question, listed in order. e.g., <tool>create_twitter_post</tool> <tool>get_last_tweet</tool> -->
  </target_tools>
  <question>
    <!-- A clear, realistic user question that requires multiple tool usage. -->
  </question>
</response>
\end{minted}

Below is an example of a task generation prompt for the multi-server task synthesis. 

\begin{minted}[fontsize=\footnotesize, linenos, breaklines, breakanywhere]{markdown}
## Task
Generate a **Multi-Server Tool Use Question** based on the provided MCP Servers and their tool descriptions.

## Objective
Analyze the provided MCP Servers and their available tools, then create a realistic user question that would naturally require the use of **{NUM_TOOLS} tools from at least 2 different MCP servers** to solve completely.

## Guidelines

### Question Realism
- Create questions that represent real-world scenarios where users would need to interact with tools from multiple MCP Servers
- The question should sound natural and authentic, as if asked by someone genuinely needing to accomplish a complex task
- Consider workflows that span across different services/domains that would require multiple servers
- Think about how different MCP servers complement each other in real-world use cases

### Server and Tool Selection
- Use tools from **at least 2 different MCP servers** to answer the question
- Select **{NUM_TOOLS} tools total** that work together across multiple servers
- The question should require a sequence or combination of tool calls from different servers to solve completely
- Choose tools based on how they complement each other across different services/domains
- Consider each tool's description and purpose when crafting the cross-server workflow
- Ensure tools from different servers create a logical, interconnected workflow

### Question Complexity
- Create questions that are complex enough to warrant using {NUM_TOOLS} tools across multiple servers
- The question should have multiple components or require several steps that span different services
- Include relevant context or constraints that make the multi-server tool usage necessary
- Do not contain the exact tool names or server names in the question
- Ensure the question cannot be reasonably answered with tools from just a single server
- Create scenarios that naturally require different types of services working together

### Cross-Server Integration
- Think about how different servers' capabilities can be combined
- Consider data flow between different services (e.g., retrieving data from one service to use in another)
- Create realistic scenarios where multiple services need to work together
- Focus on complementary functionalities across different domains

### Output Format
Your response should include:
1. **Server Analysis**: Briefly analyze all MCP Servers and their available tools, focusing on how they can work together.
2. **Cross-Server Workflow**: Describe the workflow showing how tools from different servers will be used together.
3. **Target Tools**: The specific tool names from different MCP Servers that should be used together, in the order they would likely be called, with their server names.
4. **Question**: A clear, realistic user question that requires multi-server tool usage.

## Available MCP Servers

{SERVER_DESCRIPTIONS}

## Output
Ensure your question requires exactly {NUM_TOOLS} tools from at least 2 different servers to solve completely. Provide your response in the following XML format:

<response>
  <server_analysis>
    <!-- Briefly analyze all MCP Servers and their available tools, focusing on how they can work together across different domains/services. -->
  </server_analysis>
  <cross_server_workflow>
    <!-- Describe the workflow showing how tools from different servers will be used together to solve the question. -->
  </cross_server_workflow>
  <target_tools>
    <!-- The specific tool names from different MCP Servers that should be used together, listed in order with their server names. e.g., <tool server="Server1">search_posts</tool> <tool server="Server2">send_email</tool> -->
  </target_tools>
  <question>
    <!-- A clear, realistic user question that requires multi-server tool usage spanning different services/domains. -->
  </question>
</response> 
\end{minted}

Below is an example of a task generation prompt for the task synthesis for featured servers.

\begin{minted}[fontsize=\footnotesize, linenos, breaklines, breakanywhere]{markdown}
## Task
Generate a **Multi-Server Tool Use Question** based on featured MCP Servers and their tool descriptions.

## Objective
Brainstorm a compelling real-world scenario, then analyze the provided featured MCP Servers and their available tools to create a realistic user question that would naturally require the use of **{NUM_TOOLS} tools from at least 2 different MCP servers** to solve completely.

## Guidelines

### Scenario Brainstorming
- Think of realistic, specific scenarios where someone would need to use {NUM_TOOLS} different tools across multiple servers to accomplish a meaningful task
- Consider diverse real-world contexts such as:
  - Content creators managing their online presence across different platforms
  - Researchers gathering and analyzing information from multiple sources  
  - Developers building and deploying applications using different services
  - Business professionals managing projects and communications across platforms
  - Students working on complex assignments requiring multiple tools
  - Entrepreneurs launching new ventures using various services
- The scenario should be detailed and authentic, representing genuine use cases that span multiple services

### Question Realism
- Create questions that represent real-world scenarios where users would genuinely need tools from multiple MCP servers
- The question should sound natural and authentic, as if asked by someone with a specific goal
- Include relevant context, constraints, and details that make the question engaging
- Consider workflows that require multiple complementary tools working together across different services
- Think about how different servers support each other in real-world use cases

### Server and Tool Selection
- Use tools from **at least 2 different MCP servers** to answer the question
- Select **{NUM_TOOLS} tools total** that work together across multiple servers
- The question should require a sequence or combination of tool calls from different servers to solve completely
- Choose tools based on how they complement each other across different services/domains
- Consider each tool's description and purpose when crafting the cross-server workflow
- Ensure tools from different servers create a logical, interconnected workflow

### Question Complexity
- Create questions that are complex enough to warrant using {NUM_TOOLS} tools across multiple servers
- The question should have multiple components or require several steps that span different services
- Include relevant context or constraints that make the multi-server tool usage necessary
- Do not contain the exact tool names or server names in the question
- Ensure the question cannot be reasonably answered with tools from just a single server
- Create scenarios that naturally require different types of services working together

### Cross-Server Integration
- Think about how different servers' capabilities can be combined
- Consider data flow between different services (e.g., retrieving data from one service to use in another)
- Create realistic scenarios where multiple services need to work together
- Focus on complementary functionalities across different domains

### Output Format
Your response should include:
1. **Server Analysis**: Briefly analyze the featured MCP Servers and their available tools, focusing on how they can work together.
2. **Cross-Server Workflow**: Describe the workflow showing how tools from different servers will be used together.
3. **Target Tools**: The specific tool names from different MCP Servers that should be used together, in the order they would likely be called, with their server names.
4. **Question**: A clear, realistic user question that requires multi-server tool usage.

## Available Featured MCP Servers

{FEATURED_SERVER_DESCRIPTIONS}

## Output
Ensure your question requires exactly {NUM_TOOLS} tools from at least 2 different servers to solve completely. Provide your response in the following XML format:

<response>
  <server_analysis>
    <!-- Briefly analyze the featured MCP Servers and their available tools, focusing on how they can work together across different domains/services. -->
  </server_analysis>
  <cross_server_workflow>
    <!-- Describe the workflow showing how tools from different servers will be used together to solve the question. -->
  </cross_server_workflow>
  <target_tools>
    <!-- The specific tool names from different MCP Servers that should be used together, listed in order with their server names. e.g., <tool server="Server1">search_posts</tool> <tool server="Server2">send_email</tool> -->
  </target_tools>
  <question>
    <!-- A clear, realistic user question that requires multi-server tool usage spanning different services/domains. -->
  </question>
</response> 
\end{minted}

\subsection{Task Diversification Prompt}
\label{app:task-diversification-prompts}

The following prompt aims to add diversity to the given task by introducing new contexts and personas.

\begin{minted}[fontsize=\footnotesize, linenos, breaklines, breakanywhere]{markdown}
## Task
Generate **augmented variations** of a given question that maintain the same target tool(s) usage and complexity level but apply them across different contexts and scenarios.

## Objective
Take an existing question and its associated target tool(s), then create multiple variations that:
- Use the same target tool(s) to achieve the core goal
- Maintain the exact same tool usage order and final outcome
- Apply the question to completely different contexts, scenarios, or domains
- Keep the same level of complexity and constraints as the original
- Demonstrate how the same tool usage pattern applies across diverse real-world scenarios

## Guidelines
- Translate the question to distinctly different domains, user personas, or situational contexts while preserving its original complexity level.
- Keep the tool usage sequence and final outcome identical across all variations.
- Ensure each variation feels like a realistic scenario in its new context and remains solvable with the same tool operations.
- Ensure the question does not contain any tool names or explicit references to the target tools.

## Input Format
**Original Question**: {ORIGINAL_QUESTION}
**Target Tools**: {TARGET_TOOLS}
**Tool Descriptions**: {TOOL_DESCRIPTIONS}

## Output Requirements
Generate **{VARIATIONS_COUNT} augmented variations** of the original question. Each variation should:
1. Maintain the same core goal that requires the target tool(s)
2. Use the exact same tool(s) in the same order with the same final outcome
3. Apply to a completely different context, scenario, or domain
4. Keep the same complexity level and constraints as the original
5. Feel like a natural, real-world scenario from a different setting
6. Be meaningfully different from the original and other variations in terms of context only
7. Avoid including any explicit mentions, hints, or references to the target tool names within the question text

## Output
Provide your response in the following XML format:

<response>
  <analysis>
    <!-- Briefly analyze the original question and target tool(s) to understand the core goal, tool usage pattern, complexity level, and expected outcome, then identify how this can be applied across different domains while maintaining operational consistency -->
  </analysis>
  <variations>
    <!-- Generate {VARIATIONS_COUNT} variations, each with <variation_X>, <context>, and <question> tags -->
    <variation_1>
      <context>
        <!-- Brief description of the new domain/scenario introduced -->
      </context>
      <question>
        <!-- The augmented question that maintains the same target tool(s) usage order, complexity, and outcome but in a different context -->
      </question>
    </variation_1>
    <!-- Continue with variation_2, variation_3, etc. as needed based on number of variations -->
  </variations>
</response>
\end{minted}

The prompt below is designed to enhance task complexity through the introduction of additional constraints.

\begin{minted}[fontsize=\footnotesize, linenos, breaklines, breakanywhere]{markdown}
## Task
Generate **augmented variations** of a given question that maintain the same target tool(s) usage and context but significantly increase the complexity and constraints required to solve the problem.

## Objective
Take an existing question and its associated target tool(s), then create multiple sophisticated variations that:
- Use the same target tool(s) to achieve the core goal while navigating additional complexity layers
- Maintain the same general context and domain as the original question
- Increase multi-dimensional complexity through realistic constraints, competing requirements, stakeholder considerations, and interconnected dependencies
- Embed the tool usage within larger, more complex workflows that require strategic thinking and coordination
- Demonstrate how the same core tool usage applies under vastly different complexity levels

## Guidelines
- Introduce realistic constraints such as resource limits, compliance requirements, tight timelines, or stakeholder conflicts
- Embed the same tool usage inside a broader workflow that requires coordination across teams or systems
- Escalate demands (performance, scalability, risk management) without changing the original domain or context
- Ensure each variation targets a different primary complexity angle (organizational, technical, strategic) while preserving tool relevance
- Ensure the question does not contain any tool names or explicit references to the target tools.

## Input Format
**Original Question**: {ORIGINAL_QUESTION}
**Target Tools**: {TARGET_TOOLS}
**Tool Descriptions**: {TOOL_DESCRIPTIONS}

## Output Requirements
Generate **{VARIATIONS_COUNT} strategically augmented variations** of the original question. Each variation should:
1. Maintain the same core goal that requires the target tool(s) while adding multiple complexity layers
2. Keep the same general context and domain as the original question
3. Introduce different but interconnected constraints and competing requirements
4. Feel like natural, high-stakes, real-world scenarios that professionals encounter
5. Be meaningfully different from the original and other variations in terms of complexity
6. Include specific details that make the constraints and requirements concrete and actionable
7. **Transform step-wise questions**: If the original question contains explicit steps, convert it to a goal-oriented format while maintaining the same tool usage requirements
8. Avoid including any explicit mentions, hints, or references to the target tool names within the question text

## Output
Provide your response in the following XML format:

<response>
  <analysis>
    <!-- Analyze the original question and target tool(s) to understand the core goal, current complexity level, and identify multiple complexity dimensions that can be naturally introduced while maintaining tool relevance and solution feasibility -->
  </analysis>
  <variations>
    <!-- Generate {VARIATIONS_COUNT} variations, each with <variation_X>, <constraints>, and <question> tags -->
    <variation_1>
      <constraints>
        <!-- Specific organizational, stakeholder, or coordination constraints that add realistic complexity -->
      </constraints>
      <question>
        <!-- The complex, organizationally-focused question that maintains the same target tool(s) usage within a more sophisticated workflow -->
      </question>
    </variation_1>
    <!-- Continue with variation_2, variation_3, etc. as needed based on number of variations -->
  </variations>
</response>

\end{minted}

\subsection{Task Quality Annotation Prompt}
\label{app:task-annotation-prompt-for-single-server}
\begin{minted}[fontsize=\footnotesize, linenos, breaklines, breakanywhere]{markdown}
## Task
Conduct a **Question Quality Assessment** of a tool use question across six key dimensions to ensure it meets high standards for realistic tool usage scenarios.

## Objective
Analyze the provided tool use question and assess its quality across six primary dimensions:
1. **Tool Selection Difficulty** - How challenging it is to determine which tools to use giving all available tools
2. **Tool Selection Uniqueness** - How unique and necessary the selected tools are for this specific task giving all available tools
3. **Question Quality** - Overall clarity, specificity, and effectiveness 
4. **Scenario Realism** - How authentic and believable the scenario is
5. **Verifiable** - How easy it is to verify the correctness of the final model answer
6. **Stability** - How stable the answer will be when requested under different time and geolocation

## Assessment Criteria

### 1. Tool Selection Difficulty
**What to Evaluate**: How difficult it would be for a user to determine which specific tools are needed to solve this question.

**Rating Guidelines**:
- **very easy**: Question explicitly mentions tool names or makes tool selection obvious
- **easy**: Tool selection is straightforward with clear indicators
- **medium**: Requires some reasoning but tool needs are fairly apparent  
- **hard**: Requires careful analysis to determine appropriate tools
- **very hard**: Requires extensive expertise and deep reasoning to identify the correct tools

### 2. Tool Selection Uniqueness
**What to Evaluate**: How unique and necessary the selected tools are for accomplishing this specific task, and whether the task can only be completed with these tools in the specified sequence.

**Rating Guidelines**:
- **not unique**: Many alternative tool combinations could accomplish the same task equally well
- **somewhat unique**: Some alternative approaches exist, but selected tools offer advantages
- **moderately unique**: Selected tools are well-suited, with limited alternative approaches
- **quite unique**: Selected tools are particularly well-matched to the task requirements
- **highly unique**: Task can only be accomplished effectively with these specific tools in this sequence

### 3. Question Quality
**What to Evaluate**: Overall quality, clarity, and effectiveness of the question as a realistic user query.

**Rating Guidelines**:
- **very poor**: Unclear, ambiguous, or poorly constructed question
- **poor**: Some clarity issues, missing important context
- **average**: Clear and understandable, but could be more specific or engaging
- **good**: Well-constructed, clear, specific, and realistic
- **excellent**: Exceptionally clear, detailed, engaging, and professionally written

### 4. Scenario Realism
**What to Evaluate**: How authentic, believable, and true-to-life the described scenario is.

**Rating Guidelines**:
- **unrealistic**: Artificial, contrived, or implausible scenario
- **somewhat unrealistic**: Some realistic elements but feels forced or unlikely
- **moderately realistic**: Believable scenario with minor authenticity issues
- **realistic**: Authentic scenario that represents genuine use cases
- **highly realistic**: Completely natural, authentic scenario indistinguishable from real user needs

### 5. Verifiable
**What to Evaluate**: How easy it is to verify the correctness of the final model answer.

**Rating Guidelines**:
- **hard to verify**: Fully free-form answer that requires extensive human judgment
- **somewhat hard**: Mostly subjective answer with some verifiable elements
- **moderately verifiable**: Short sentence that can be verified by LLM comparison
- **mostly verifiable**: Answer with clear, objective components and some subjective elements
- **easy to verify**: Answer can be verified by simple rules, exact matches, or clear success criteria

### 6. Stability (1-5 Scale)
**What to Evaluate**: How stable and consistent the answer will be when the question is asked under different environmental conditions and system contexts. Consider factors like temporal dependency, geographical variations, operating system differences, network environments, and software version variations.

**Rating Guidelines**:
- **highly unstable**: Answer changes significantly across different conditions (real-time data, location-specific, system-dependent)
- **somewhat unstable**: Answer may vary moderately based on environmental or system factors
- **moderately stable**: Answer mostly consistent with minor variations due to context
- **mostly stable**: Answer remains largely consistent across different conditions
- **highly stable**: Answer is completely independent of environmental and system factors

## Question Analysis

### All Available Tools```
{ALL_SERVER_AND_TOOL_INFORMATION}
```

### Question Content
```
{QUESTION_CONTENT}
```

### Intended Tool for This Question
```
{INTENDED_TOOL}
```

## Output Requirements

Provide analysis with detailed reasoning BEFORE scores for each of the six metrics.

## Output
Provide your response in the following XML format:

<response>
  <tool_selection_difficulty>
    <reasoning>
      <!-- Detailed explanation including ambiguity level, domain knowledge required, and alternative solutions giving all available tools -->
    </reasoning>
    <rating><!-- Rating: very easy, easy, medium, hard, very hard --></rating>
  </tool_selection_difficulty>
  
  <tool_selection_uniqueness>
    <reasoning>
      <!-- Detailed explanation of tool necessity, sequential dependencies, and alternative tool viability giving all available tools -->
    </reasoning>
    <rating><!-- Rating: not unique, somewhat unique, moderately unique, quite unique, highly unique --></rating>
  </tool_selection_uniqueness>
  
  <question_quality>
    <reasoning>
      <!-- Detailed explanation covering linguistic quality, information architecture, and actionability -->
    </reasoning>
    <rating><!-- Rating: very poor, poor, average, good, excellent --></rating>
  </question_quality>
  
  <scenario_realism>
    <reasoning>
      <!-- Detailed explanation of industry authenticity, workflow accuracy, and stakeholder behavior -->
    </reasoning>
    <rating><!-- Rating: unrealistic, somewhat unrealistic, moderately realistic, realistic, highly realistic --></rating>
  </scenario_realism>
  
  <verifiable>
    <reasoning>
      <!-- Detailed explanation of answer format, objective criteria, and ground truth availability -->
    </reasoning>
    <rating><!-- Rating: hard to verify, somewhat hard, moderately verifiable, mostly verifiable, easy to verify --></rating>
  </verifiable>
  
  <stability>
    <reasoning>
      <!-- Detailed explanation of temporal/geographical/system dependencies and environmental factors -->
    </reasoning>
    <rating><!-- Rating: highly unstable, somewhat unstable, moderately stable, mostly stable, highly stable --></rating>
  </stability>
</response> 

\end{minted}

\subsection{Trajectory Annotation Prompt}
\label{app:trajectory-annotation-prompt-for-single-server}
\begin{minted}[fontsize=\footnotesize, linenos, breaklines, breakanywhere]{markdown}
## Task
Conduct a **Response Quality Assessment** of a tool-use conversation across two LLM-scored dimensions, with a third dimension computed automatically outside the LLM.

## Objective
Analyze the provided conversation and assess its response quality across two primary dimensions scored by the LLM, while reserving an additional tool-call accuracy dimension for automated scoring:
1. Completeness - Whether the assistant fully accomplished the user's request end-to-end
2. Conciseness - Whether the assistant solved the task using the minimum necessary steps and verbosity

## Assessment Criteria

### 1. Completeness
**What to Evaluate**: Did the assistant fully satisfy the user's goal given the conversation context? Consider whether the assistant:
- Executed all required steps end-to-end (including saving/exporting/downloading where applicable)
- Provided the final deliverable or a working alternative when blocked (e.g., tool failure with a usable fallback)
- Included essential confirmations, paths, or instructions to achieve the outcome
- Avoided missing key requirements or leaving the user with unresolved gaps

**Rating Guidelines**:
- very incomplete: Major requirements missing; no usable outcome
- incomplete: Some key requirements missing; outcome is not directly usable
- partially complete: Core steps attempted; outcome usable only with user effort or missing minor requirements
- mostly complete: Meets most requirements; small omissions or minor issues remain
- fully complete: All requirements met with a usable outcome delivered

### 2. Conciseness
**What to Evaluate**: Did the assistant achieve the goal with minimal redundancy and steps? Consider whether the assistant:
- Avoided repetitive or unnecessary explanations/tool calls
- Used the minimal set of steps/tools to complete the task
- Kept language concise while preserving clarity

**Rating Guidelines**:
- very redundant: Excessive repetition or unnecessary steps/tool calls
- redundant: Noticeable verbosity or extra steps beyond what's needed
- average: Reasonably concise with minor extraneous content
- concise: Efficient and to the point with minimal overhead
- very concise: Maximally efficient while clear and complete

## Response Analysis

### Question Content
```
{QUESTION_CONTENT}
```

### Intended Tool for This Question
```
{INTENDED_TOOL}
```

### Conversation History
```
{CONVERSATION_HISTORY}
```

## Output Requirements
- Provide detailed reasoning BEFORE ratings for Completeness and Conciseness
- Do NOT score Tool Call Accuracy; include placeholders only

## Output
Provide your response in the following XML format:

<response>
  <completeness>
    <reasoning>
      <!-- Evaluate if the assistant delivered an end-to-end usable outcome, addressed all requirements, handled tool failures with alternatives, and provided necessary confirmations/paths. -->
    </reasoning>
    <rating><!-- Rating: very incomplete, incomplete, partially complete, mostly complete, fully complete --></rating>
  </completeness>

  <conciseness>
    <reasoning>
      <!-- Evaluate if the assistant minimized redundant steps/explanations, avoided unnecessary tool calls, and kept messaging efficient while clear. -->
    </reasoning>
    <rating><!-- Rating: very redundant, redundant, average, concise, very concise --></rating>
  </conciseness>
</response>
\end{minted}

% \subsection{Assets}
% \begin{itemize}
%     \item checkpoints internal: /proj/checkpoints/zhangchen/tool-rl-dev/sft_models
%     \item All data internal: /proj/checkpoints/zhangchen/tool-rl-dev/data
%     \item Offial Release: /proj/checkpoints/zhangchen/tool-rl-dev/data/7.Final_Release \/ we release raw (after rule-based trajectory filtering) and filtered (high quality subsset, after llm annotation) \/train: after applying ms-sw format for sft (so same as filtered)
%     \item Raw Data: (1) https://huggingface.co/datasets/zhangchenxu/WildMCP-KimiK2 (2) https://huggingface.co/datasets/zhangchenxu/WildMCP-OSS (3) https://huggingface.co/datasets/zhangchenxu/WildMCP-Qwen3
%     \item All public HF assets: models, public datasets: HF org name focused on tool and agent!!!
%     \item GitHub repo:
%         Private: \url{https://github.com/zhangchen-xu/tool-rl-dev}
%         Public: TBD once we have the name
%     \item benchmark results spreadsheet: \url{https://drive.google.com/drive/folders/1-8gO3oPLrjfImqkHjntqupK8_2Wg2GV_?usp=drive_link}
    
% \end{itemize}
\end{document}